\pdfoutput=1
\documentclass{article}
\PassOptionsToPackage{dvipsnames,table,xcdraw}{xcolor}

\usepackage{balance}
\usepackage{microtype}
\usepackage{graphicx}
\usepackage{subfigure}
\usepackage{booktabs} 
\usepackage{setspace}
\usepackage{hyperref}
\usepackage{longtable}

\usepackage[accepted]{icml2025}

\usepackage{amsmath}
\usepackage{amssymb}
\usepackage{mathtools}
\usepackage{amsthm}

\usepackage[utf8]{inputenc}  
\usepackage[T1]{fontenc}

\usepackage[capitalize,noabbrev]{cleveref}

\theoremstyle{plain}

\theoremstyle{definition}

\theoremstyle{remark}

\usepackage[textsize=tiny]{todonotes}

\usepackage{amsmath,bm}
\usepackage{enumitem}

\usepackage{colortbl}
\usepackage{adjustbox}
\usepackage{multirow}
\usepackage{pifont}
\usepackage{float}
\usepackage{xspace}
\usepackage{tcolorbox}
\usepackage{twemojis}
\usepackage[fixed]{fontawesome5}

\definecolor{maroon}{cmyk}{0,0.1,0.01,0.01}
\definecolor{blue}{cmyk}{0.95,0.0,0.2,0.2}
\definecolor{yellow}{cmyk}{0.01,0.0,0.2,0.01}
\definecolor{lightblue}{cmyk}{0.1,0.0,0.02,0.02}
\definecolor{mypink}{RGB}{219, 48, 122}
\definecolor{spink}{HTML}{E91E62}
\definecolor{spurple}{HTML}{3F51B5}

\definecolor{medalign}{HTML}{2c74dc}         
\definecolor{timer-bench-edge}{HTML}{B22222}  
\definecolor{timer-bench-uniform}{HTML}{228B22} 

\newcommand{\base}{{TIMER}\xspace}
\newcommand{\instruct}{{TIMER-Instruct}\xspace}
\newcommand{\benchmark}{{TIMER-Bench}\xspace}

\newcommand{\xmark}{\ding{55}}

\icmltitlerunning{\base: Temporal Instruction Modeling and Evaluation for Longitudinal Clinical Records}

\begin{document}

\twocolumn[
\icmltitle{\base: Temporal Instruction Modeling and Evaluation for \\Longitudinal Clinical Records}



\icmlsetsymbol{equal}{*}
\icmlsetsymbol{stanford}{\twemoji{evergreen tree}}

\begin{icmlauthorlist}
\icmlauthor{Hejie Cui}{equal,stanford}
\icmlauthor{Alyssa Unell}{equal,stanford}
\icmlauthor{Bowen Chen}{stanford}
\icmlauthor{Jason Alan Fries}{stanford}
\icmlauthor{Emily Alsentzer}{stanford} \\
\icmlauthor{Sanmi Koyejo}{stanford}
\icmlauthor{Nigam Shah}{stanford}

\icmlEqualContribution \quad\quad\quad
\twemoji{evergreen tree} Stanford University

\end{icmlauthorlist}


\icmlcorrespondingauthor{Hejie Cui}{hejie.cui@stanford.edu}
\icmlcorrespondingauthor{Alyssa Unell}{aunell@stanford.edu}
\icmlcorrespondingauthor{Nigam Shah}{nigam@stanford.edu}
\icmlkeywords{Machine Learning, ICML}

\vskip 0.3in
]



\printAffiliationsAndNotice{\icmlEqualContribution} 

\begin{abstract}
Large language models (LLMs) have emerged as promising tools for assisting in medical tasks, yet processing Electronic Health Records (EHRs) presents unique challenges due to their longitudinal nature. While LLMs' capabilities to perform medical tasks continue to improve, their ability to reason over temporal dependencies across multiple patient visits and time frames remains unexplored. 
We introduce \textbf{\base} (\textbf{T}emporal \textbf{I}nstruction \textbf{M}odeling and \textbf{E}valuation for Longitudinal Clinical \textbf{R}ecords), a framework that incorporate instruction-response pairs grounding to different parts of a patient's record as a critical dimension in both instruction evaluation and tuning for longitudinal clinical records.
We develop \benchmark, the first time-aware benchmark that evaluates temporal reasoning capabilities over longitudinal EHRs, as well as \instruct, an instruction-tuning methodology for LLMs to learn reasoning over time. 
We demonstrate that models fine-tuned with \instruct improve performance by 7.3\% on human-generated benchmarks and 9.2\% on \benchmark, indicating that temporal instruction-tuning improves model performance for reasoning over EHR. Our code is available at \href{https://anonymous.4open.science/r/TIMER-2874}{TIMER}.

\end{abstract}

\section{Introduction}
\label{sec:intro}
While many language models now handle context lengths of hundreds of thousands of tokens, their ability to reason across longitudinal documents and follow complex instructions remains limited~\cite{li2024long,kuratov2024babilong}. This limitation is particularly critical in healthcare, where physicians routinely analyze electronic health records (EHRs) spanning multiple years and thousands of entries~\cite{huguet2020using}. Tasks such as chronic disease management, multi-visit care planning, and patient history synthesis require clinicians to understand complex relationships between different record entries and how past events influence current and future clinical decisions~\cite{wornow2024context}. The cognitive demands of processing such lengthy documentation are significant. While biomedical LLMs have shown promising results on well-structured tasks like answering USMLE questions and medical knowledge retrieval~\cite{singhal2023large,lu2024large,lucas2024reasoning}, recent evaluations reveal their significant limitations in processing longitudinal patient information and in making clinical decisions over time~\cite{hager2024evaluating,bedi2024testing}. The gap between isolated question-answering performance and temporal reasoning ability impacts the practical utility of LLMs in healthcare. While there is some prior work that has explored temporal understanding abilities of general LLMs ~\cite{wang2024trambenchmarkingtemporalreasoning, fatemi2024testtimebenchmarkevaluating, herel2024timeawarenesslargelanguage}, how these capabilities scale to longer contexts remains understudied, particularly in healthcare where longitudinal reasoning is important.


\begin{figure}[t]
\includegraphics[width=\linewidth]{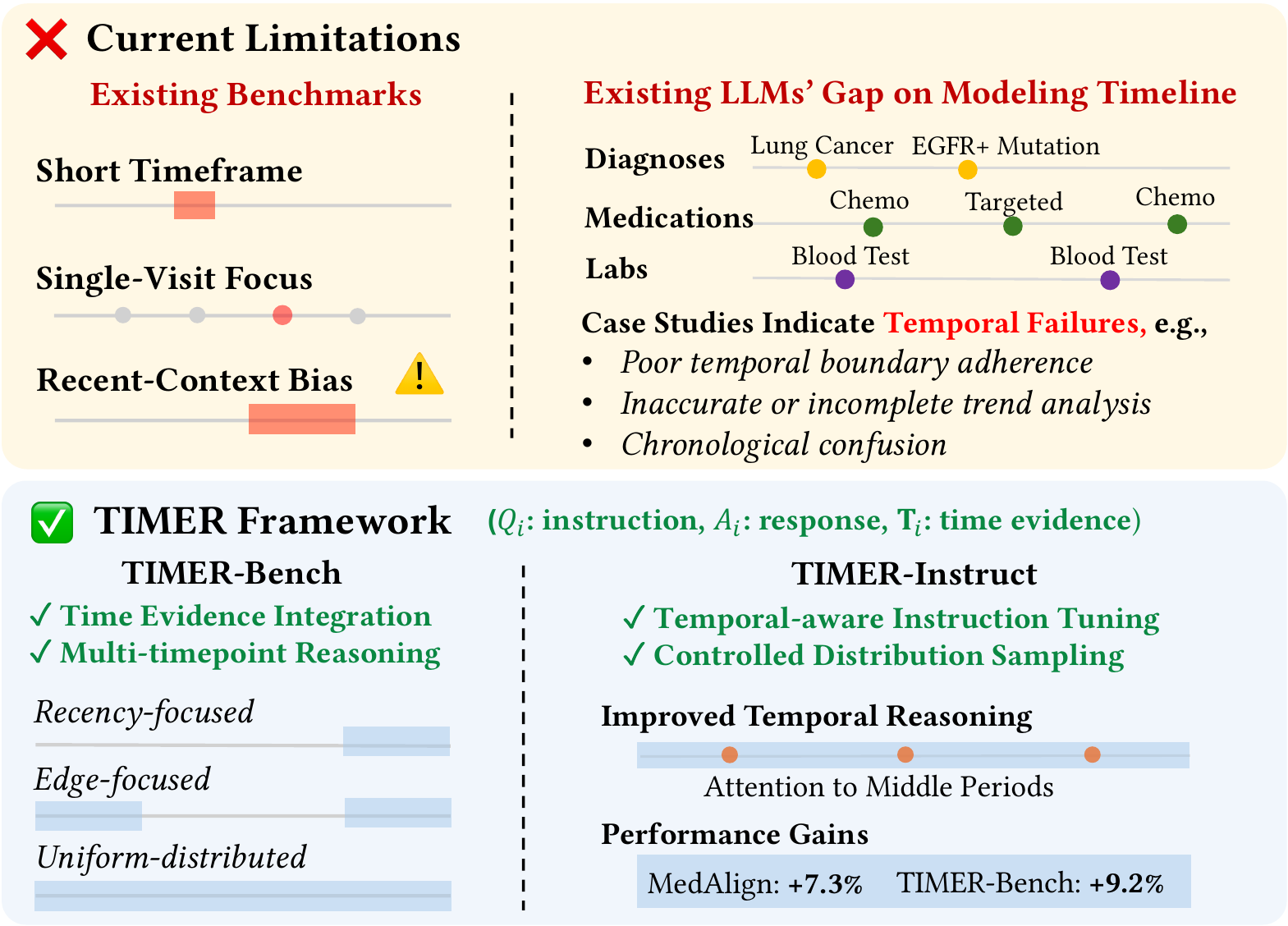} 
\vskip -1em
\caption{Overview of temporal instruction modeling challenges and our TIMER framework. Top: Existing benchmarks suffer from limited temporal coverage and recent-context bias, while baseline models show critical failures in longitudinal reasoning. Bottom: The TIMER framework addresses these limitations through temporal-aware distribution evaluation (TIMER-Bench) and instruction tuning (TIMER-Instruct), achieving significant improvements in both human-curated and model-generated benchmarks.}
\vskip -1.5em
\label{fig:tldr}
\end{figure}

Instruction tuning has been proved useful to adapt LLMs for domain-specific tasks~\cite{zhang2023instruction}. However, the temporal aspects of instruction evaluation and tuning—particularly how the temporal distribution of information affects model performance—remain poorly understood~\cite{scheller2022temporal}. As illustrated in Figure~\ref{fig:tldr} top-left, existing benchmarks often overlook the inherent temporal complexity of reasoning over medical records~\cite{wu2024instruction}. While some benchmarks include longer patient timelines, they don't systematically account for temporal distribution of information in their evaluation design and typically assess recall over recent notes/events from the patient's longitudinal medical record~\cite{fleming2024medalign}. This recency bias in benchmarks limits understanding of how models handle information across different time points in patient timelines~\cite{zhao2019learning,paik2021condensed}.

To address these challenges, we introduce \benchmark, an evaluation benchmark designed to evaluate temporal reasoning over longitudinal patient records. Given the challenges of human curation for lengthy records, we leverage LLMs to generate instruction-response pairs with \textit{explicit temporal evidence}.  \benchmark enables instruction sampling with different temporal distributions, allowing systematic evaluation across patient timelines as shown in Figure~\ref{fig:tldr} bottom-left. \benchmark includes queries that require reasoning over multiple time points, providing an assessment of the model's ability to synthesize information from varying parts of the patient timeline. 
However, evaluation of existing LLMs on \benchmark shows several failure modes such as poor temporal boundary adherence, where models include irrelevant historical data; inaccurate trends, where models fail to track pattern over time; and chronological confusion, where models mix up the order of events.

To improve the longitudinal reasoning capability of LLMs, we introduce \instruct, a methodology for temporal instruction tuning using model-generated instruction-response pairs that ground to different parts of EHR. 
We identify that model-generated data exhibit a ``lost-in-the-middle'' pattern by default—focusing on timeline extremes while overlooking middle periods. Therefore, we investigate different instruction temporal distributions that focus on controlled subsets of patient visits: recency-focused instructions, edge-focused instructions, and uniformly distributed instructions. Our evaluation shows that while distribution alignment between training and evaluation generally improves performance, uniformly distributed instructions show particular strength on uniformly distributed \benchmark where questions spread mid-timeline reasoning. This result demonstrates the importance of a balanced temporal coverage of the instruction-response pairs during instruction tuning when downstream tasks involve reasoning over the full temporal range. Compared to baselines without temporal instruction tuning, models instruction-tuned with \instruct improve performance by 7.3\% on MedAlign and 9.2\% on temporal reasoning benchmark \benchmark.

In summary, this work makes three key contributions: (1) we identify temporal distribution of instructions as a critical yet overlooked dimension in the evaluation of clinical language models, demonstrating that the temporal gaps present in existing benchmarks limit our understanding of the true reasoning capabilities of LLM models; (2) building on this insight, we introduce a new benchmark \benchmark to assess longitudinal reasoning capabilities. Our findings reveal that models struggle to reason over time in a longitudinal context; and (3) We develop \instruct, a new methodology for instruction tuning which takes temporal dependencies into account, achieving state-of-the-art performance on physician-generated benchmarks and \benchmark. 
To facilitate further research, we make our code available at \href{https://anonymous.4open.science/r/TIMER-2874}{TIMER}. TIMER-Bench will be released under a research data use agreement to support responsible evaluation.

\section{Related Work} 
\label{sec:background}

\subsection{LLM Evaluation on Longitudinal Clinical Records}
Electronic Health Records (EHRs) serve as comprehensive digital repositories of patient care, containing structured data (diagnosis codes, medications, lab results), unstructured clinical notes, demographic information, and date time across visits~\cite{theodorou2023synthesize}. To provide effective care, physicians must synthesize this complex data to track disease progression, treatment responses, and temporal relationships between medical events across different providers \cite{carrasco2023prediction,allam2021analyzing}.
With the emergence of LLMs in healthcare, evaluation approaches have evolved from testing medical knowledge through standardized exams like USMLE and clinical vignettes~\cite{singhal2023large,goh2024influence} to more complex tasks such as diagnosis prediction~\cite{hager2024evaluating}, medication recommendation~\cite{inoue2024drugagent}, and clinical note summarization~\cite{van2024adapted,tang2023evaluating}. However, evaluating models on longitudinal records introduces distinct challenges, including ensuring both medical accuracy and temporal understanding, handling lengthy multi-visit documentation, and the high costs of expert annotation at scale~\cite{seo2024evaluation,zhu2024prompting}.

\subsection{Instruction Tuning for Medical Applications}
With the emergence of foundation models and their capability to perform various tasks, instruction tuning has served as a crucial method to align these models with user preferences~\cite{ouyang2022training,zhang2023instruction}. While effective, implementation is often constrained by the limited availability of high-quality instruction-response pairs. This has driven interest in synthetic data generation~\cite{dubois2024alpacafarm,wang2023self}. However, the role of synthetic data in model training remains contentious and concerns persist about model collapse and degraded performance~\cite{kazdan2024collapse}. Recent research emphasizes instruction diversity for robust model alignment~\cite{ge2024scaling,cui2024biomedical}. Nonetheless, these approaches face a critical limitation: the inability of models to synthesize information from long contexts effectively. 
This challenge is particularly relevant in medical applications, where a comprehensive understanding of longitudinal patient timelines is often crucial for accurate decision-making. Initial medical instruction tuning approaches mainly focus on brief instruction responses, often reduced to simple retrieval tasks or fact-based question-answering~\cite{zhang2023alpacare,tran2024bioinstruct,rohanian2024exploring}, which fails to capture the complex reasoning required in real clinical settings.

\section{Temporal Evaluation Gaps in EHR Instruction Benchmarks}
\label{sec:limitation}

\textbf{Limitation 1: Single visit and short temporal coverage in ICU-based instructions.}
The existing EHR instruction data primarily stems from MIMIC records, which is a standard collection of ICU patient timelines. Consequently, MIMIC data does not contain longitudinal timelines beyond ICU visits, resulting in this data lacking complex temporal relationships that span multiple visits from varying time frames. The MIMIC-Instr dataset~\cite{wu2024instruction} comprises instruction response pairs derived from the discharge summary in MIMIC-IV for patients admitted to ICUs, and those pairs were generated from single visits, with an average duration of just 7.2 days (171.8 hours). This limitation arises directly from the source data, which is heavily skewed toward single-visit patients, comprising 84.5\% (n=3,639) of the dataset, whereas multi-visit patients make up only 15.5\% (n=670). Additionally, not all notes from a single visit are included, as much of a patient's stay occurs on the floor instead of in the ICU. This narrow temporal perspective fails to address the complexities involved in chronic disease management, long-term treatment outcomes, and the progression of patient conditions over extended periods.



\begin{figure}[t]
\includegraphics[width=\linewidth]{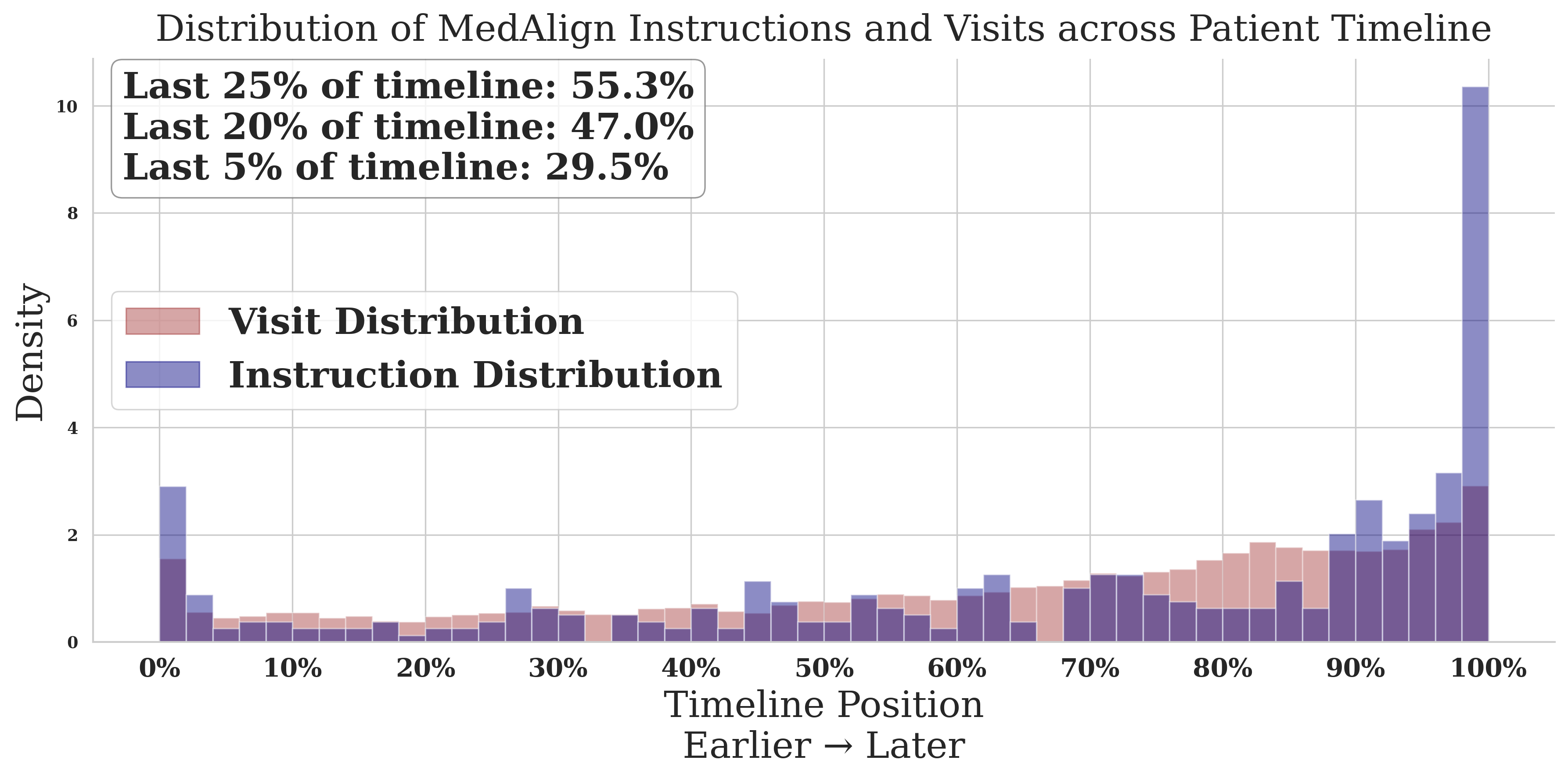} 
\vskip -1em
\caption{MedAlign instruction benchmark for longitudinal records emphasizes recent portions of each patient's longitudinal record.}
\vskip -1em
\label{fig:medalign-distribution-bias}
\end{figure}

\textbf{Limitation 2: Recency bias in physician benchmarks.}
MedAlign~\cite{fleming2024medalign} is introduced as the first benchmark created by clinicians covering realistic clinical instructions. However, physician-curated benchmarks show a natural bias toward recent records, as reviewing long patient histories is cognitively demanding, and creating instruction data manually from scratch is time-consuming. Despite covering a substantial average timespan of 3,895.06 days (approximately 10.7 years), the distribution of clinical instructions is concentrated on recent patient visits. Figure~\ref{fig:medalign-distribution-bias} illustrates the distribution of physician-created instruction and the actual patient visits across timelines. The time evidence information is extracted from human-written rationales accompanying the human-written responses in the dataset. It shows that 55.3\% of all clinical instructions (n=396) are concentrated in just the last quarter of the patient timeline, with an even more pronounced 47.0\% in the last 15\% and 29.5\% in the final 5\%. While this partly reflects clinical practice—where recent summaries capture trajectory information—it may overlook crucial reasoning during earlier periods. Moreover, 71.3\% of questions in MedAlign are retrieval-based, focusing on ``needle-in-the-haystack'' capabilities rather than evaluating the model's ability to synthesize over the full patient timeline. 

\textbf{Motivation.} These temporal gaps in existing EHR instruction evaluation benchmarks – either constrained to brief ICU stays or biased toward recent events in longer records – highlight a critical gap in evaluation. Temporal bias in instruction distribution across patient records, along with insufficient coverage of longitudinal data, impedes the development of models capable of reasoning about long-term patient trajectories. To address these gaps, we aim to develop instruction benchmarks with explicit time evidence provenance spanning years of clinical records. This enables evaluating models' capabilities across comprehensive patient histories with controlled temporal distributions, moving beyond the constraints of acute episodes or recent events.

\section{\base}
\label{sec:method}
\begin{figure*}[t]
\centering
  \includegraphics[width=\textwidth]{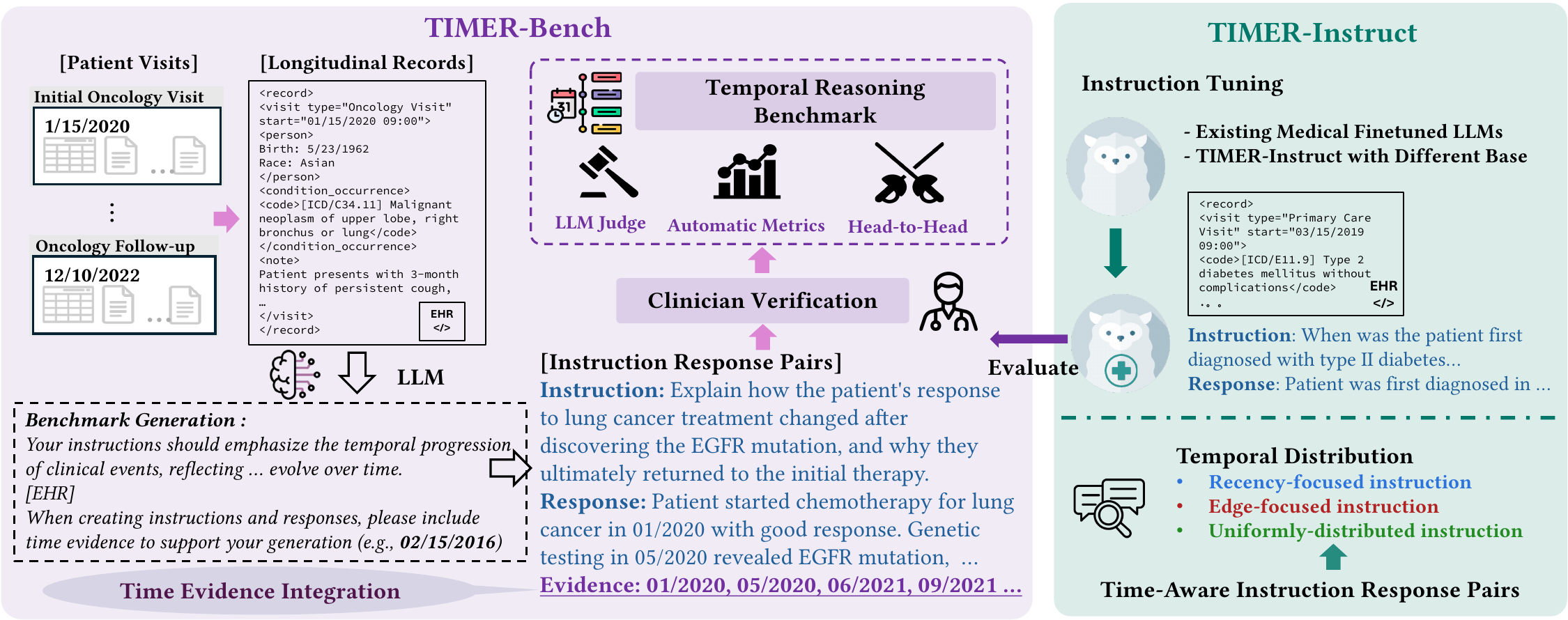}
    \vskip -1em
    \caption{Overview of \textbf{\base} framework. \benchmark creates evaluation sets with explicit temporal evidence, covering questions across different time periods in patient histories to assess longitudinal EHR reasoning. Right: \instruct enhances model performance through instruction tuning with instruction-response pairs generated by LLMs that distribute temporally diverse across EHR timelines.}
    \vskip -1em
    \label{fig:framework}
\end{figure*}

We introduce \base, a framework for evaluating and enhancing temporal reasoning capabilities of LLMs on longitudinal EHRs, as present in Figure~\ref{fig:framework}. \base consists of two components: (1) \benchmark, which generates evaluation sets with explicit time evidence integration, and (2) \instruct, which improves models' longitudinal reasoning through temporal-aware instruction tuning. 

\subsection{\benchmark: An Evaluation Benchmark for LLM Longitudinal Reasoning }
\label{ssec:longehr-bench}

\textbf{Multiple-timepoint reasoning evaluation.}
EHRs present unique challenges for LLM evaluation due to their inherently temporal and longitudinal nature. Unlike general long text documents, medical records capture a patient's healthcare journey through time, where each clinical visit is intrinsically linked to when it occurred and how it relates to other visits in the patient's history. Complex clinical reasoning often requires synthesizing information across multiple time points. For example, consider an instruction: ``\textit{Explain how the patient's response to lung cancer treatment changed after discovering the EGFR mutation, and why they ultimately returned to the initial therapy.}'' with patient records: 
\begin{tcolorbox}[
colback=cyan!5!gray!5,
colframe=cyan!30!gray!30,
left=5pt,
boxsep=2pt
]
\begin{itemize}[nosep,leftmargin=*]
    \item[\textcolor{blue!70!black}{$\blacktriangleright$}] \textbf{01/2020}: initial diagnosis of lung cancer and started chemotherapy with good response
    \item[\textcolor{blue!70!black}{$\blacktriangleright$}] \textbf{05/2020}: genetic testing in revealed EGFR mutation, switched to targeted therapy
    \item[\textcolor{blue!70!black}{$\blacktriangleright$}] \textbf{08/2020-06/2021}: clinical notes showed exceptional response with tumor shrinkage
    \item[\textcolor{blue!70!black}{$\blacktriangleright$}] \textbf{09/2021}: resistance developed with new growth
    \item[\textcolor{blue!70!black}{$\blacktriangleright$}] \textbf{01/2022-12/2022}: restarted original chemotherapy and provided excellent control
\end{itemize}
\end{tcolorbox}

This instruction illustrates how temporal reasoning requires synthesizing information across timelines. An evaluation focusing only on the beginning and end of the record might miss crucial intermediate events, such as the development of resistance to targeted therapy in \textbf{09/2021}—a critical turning point that explains the eventual return to initial therapy. Understanding such treatment trajectories requires models to attend to and integrate information from all periods of the patient history. While real clinical records contain these rich temporal patterns, human-curated instruction-response pairs for such lengthy histories are time-consuming. This challenge motivates our approach of leveraging LLMs to systematically generate temporally grounded evaluation data that require multiple time points for longitudinal reasoning.

\textbf{Benchmark generation process.}
To provide a comprehensive evaluation, we developed an approach for generating benchmarks that integrates time evidence. This method explicitly preserves and utilizes temporal relationships during the creation process, reflecting how clinicians interact with EHRs by actively considering both the content and the temporal context of medical events. As shown in the left panel of Figure~\ref{fig:framework}, we begin by aggregating patient visits and converting them into XML-formatted longitudinal records. These records are then used as input for a language model to generate instruction-response pairs. During the prompting process, we instruct the language model to provide date-time evidence \(\mathbf{T}_i = \{T_{i,1}, ..., T_{i,n_i}\}\) for each instruction-response pair \((Q_i, A_i)\) in the benchmark. This evidence connects the instance to related visits in the patient's timeline, resulting in tuples of \((Q_i, A_i, \mathbf{T}_i)\). This temporal grounding ensures that each benchmark instance captures meaningful temporal relationships while maintaining clinical accuracy. To effectively capture multi-timestamp reasoning, we provide explicit guidance for generating instances that integrate information from multiple visits. This approach encourages the referencing of specific events from different time points, ensuring that the benchmark truly evaluates longitudinal reasoning rather than just single-visit comprehension.


\textbf{Clinical validation.} 
To ensure benchmark quality, we conducted validation with three clinicians who reviewed 100 randomly sampled instruction-response pairs from \benchmark. The clinicians evaluated three key aspects: clinical relevance, which measures alignment with real-world medical scenarios; temporal reasoning complexity, which assesses the depth of temporal synthesis required; and factual accuracy, which verifies perceived medical correctness. The evaluation results included average scores of 95/100 for clinical relevance, 80/100 for temporal reasoning complexity, and 98/100 for factual accuracy. This validation process confirms that \benchmark maintains high clinical authenticity while effectively testing models' long context temporal reasoning capabilities.

\subsection{\instruct: Instruction Tuning for Longitudinal Clinical Records}
To improve model performance on temporal reasoning tasks, we develop \instruct, a methodology for instruction tuning that utilizes time-aware instruction-response pairs generated by LLMs. 

\textbf{Temporal variability in clinical records.}
A key challenge in analyzing temporal patterns across patient records is that even with fixed context lengths, the actual time spans can vary significantly due to irregular intervals between clinical events. In clinical records, events like diagnoses, medications, laboratory results, and procedures are distributed irregularly across time, reflecting the reality of patient care where visits may be clustered during acute episodes or spread out during stable periods. To understand temporal patterns in model-generated instructions, we need a way to analyze these varying timescales systematically.

\begin{figure}[t]
\includegraphics[width=\linewidth]{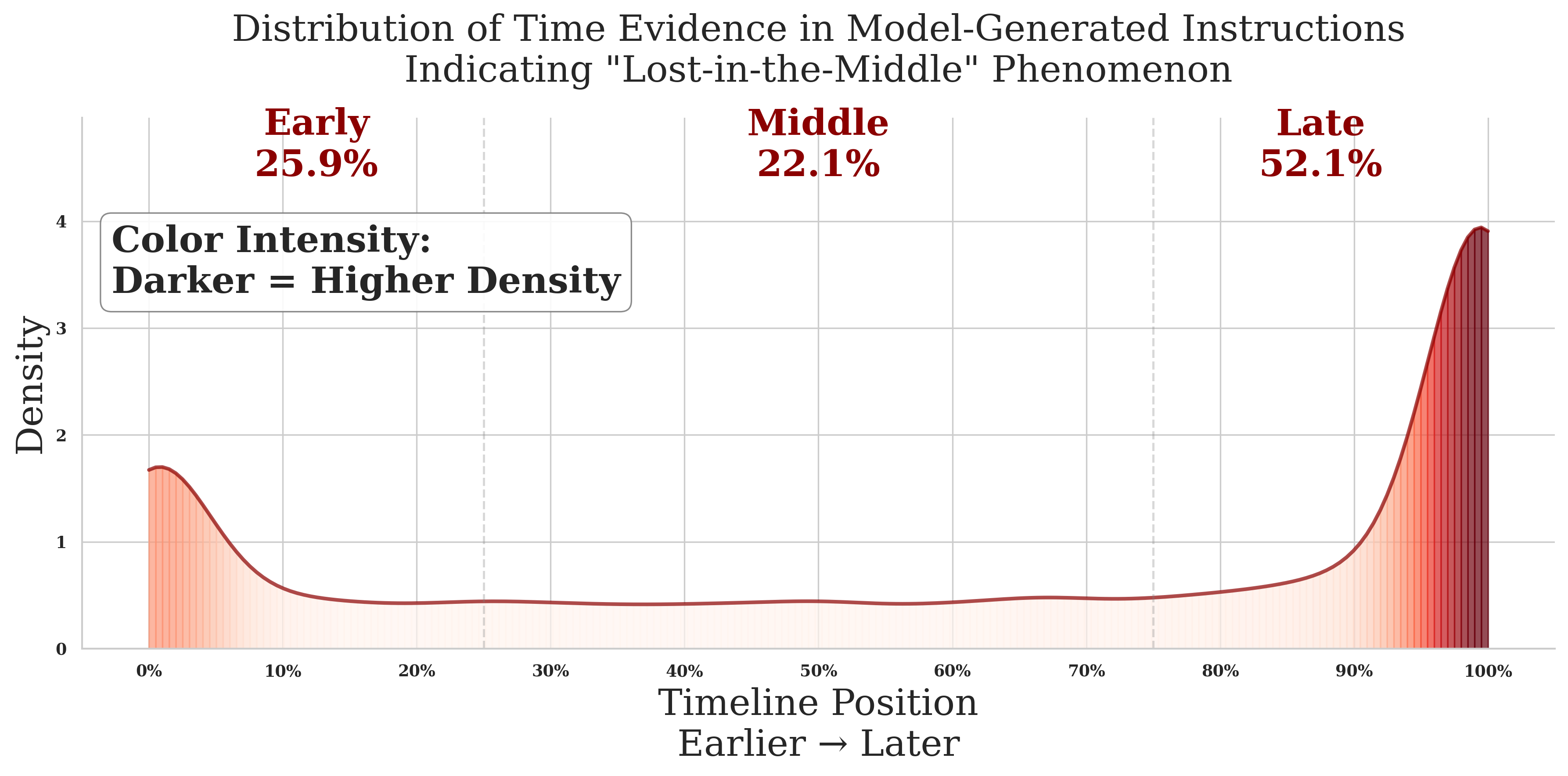} 
\centering
\vskip -1.0em
\caption{Temporal distribution in model-generated instruction-response pairs reveals a ``lost-in-the-middle" phenomenon. Using our normalized temporal position metric (x-axis: 0\% to 100\% of timeline), we find that instructions strongly favor timeline edges. The density plot shows high concentration at recent and early periods, while middle periods receive significantly less attention. }
\vskip -1.0em
\label{fig:temporal-distribution}
\end{figure}

\textbf{Temporal distribution analysis.} 
We introduce a normalized position metric to account for these varying timescales. For each time evidence $T_j \in \mathbf{T}_i$ of an instruction-response pair $(Q_i, A_i)$, we define the relative temporal position $P_j$ as:
\begin{equation}
\setlength{\abovedisplayskip}{5pt}
\setlength{\belowdisplayskip}{5pt}
P_j = \frac{T_j-T_{min}}{T_{max}-T_{min}}
\end{equation}
where $T_j$ represents an evidence timestamp of this pair, and $T_{min}$ and $T_{max}$ are date time of visits that bound the context window. This normalization enables us to compare temporal patterns across patient records with different absolute time spans while maintaining relative temporal relationships.
Using such normalized temporal metrics, we analyze the distribution of temporal evidence in model-generated instructions across patient timelines. Figure~\ref{fig:temporal-distribution} reveals a striking ``lost-in-the-middle'' phenomenon in the default generation of instruction-response pairs. 
Such edge-focused distribution indicates that when generating instruction data, LLMs tend to pay more attention to early and recent events in long contexts, while neglecting the intermediate period. These inherent biases in temporal focus during the LLM-based instruction generation could result in tuned models that overlook important developments in patient timelines.

\textbf{Instruction tuning with various temporal distribution patterns.}
Motivated by the observed ``lost-in-the-middle'' phenomenon in model-generated instruction-response pairs and the temporal biases in existing evaluation benchmarks, we explore how temporal distributions in tuning data affect model performance. We construct three instruction-tuning sets from the same set of patient longitudinal records, each reflecting a distinct distribution of instructions' relative temporal position, as demonstrated in Figure~\ref{fig:framework} right panel: 
\begin{itemize}[nosep,leftmargin=*]
\item \textit{Recency}: a recency-focused set that concentrates instructions in the last quartile of the timeline, similar to the temporal patterns observed in human-annotated datasets
\item \textit{Edge}: an edge-focused set which exhibits higher instruction density regarding visits at the context boundaries $T_{min}$ and $T_{max}$ of patient timelines, similar to the natural temporal patterns generated by large language models
\item \textit{Uniform}: a uniformly-distributed set that ensures balanced temporal coverage by maintaining consistent instruction density across all relative positions
\end{itemize}
By maintaining consistent size and patient timelines across these instruction sets and varying only their temporal distributions over the timeline, we isolate temporal positioning as the key variable of how training data patterns influence models' ability to reason across different time points.

\begin{table*}[h]
\centering
\caption{Dataset information of existing EHR instruction benchmarks and \benchmark.}
\resizebox{\linewidth}{!}{%
\begin{tabular}{l|ccccc|c}
\toprule
\bf Dataset & Multi-Visit & Avg Time Span & Raw Record Modality & Curation & Size Scalable &\bf Attribute \\
\midrule
\bf MIMIC-Instr (Test) & No & 7.2 days & Notes Only & Model-Synthetic & Yes (200) & Instruction + Response \\
\bf MedAlign & Yes & 3895.06 days & Structured Data \& Notes & Human-Created & No (303) & Instruction + Response + Rationale \\
\bf \benchmark & Yes & 1294.88 days & Structured Data \& Notes & Model-Synthetic & Yes (402) & Instruction + Response + Time Evidence \\
\bottomrule
\end{tabular}
}
\vskip -0.5em
\label{tab:benchmark-stats}
\end{table*}

\section{Experiments}

\begin{table*}[t]
\centering
\caption{Performance (\%) of baseline models and \instruct on MedAlign and \benchmark benchmarks, reported as mean ± standard deviation from bootstrap resampling (n=10,000) over the test set with 100 samples.}
\label{tab:instruct-performance}
\renewcommand{\arraystretch}{0.88}
\resizebox{\linewidth}{!}{
\begin{tabular}{p{5cm}|cccc|cccc}
\toprule
\multirow{4}{*}{\textbf{Model Name}} & \multicolumn{4}{c|}{\textbf{MedAlign}} & \multicolumn{4}{c}{\textbf{\benchmark}} \\
\cmidrule(lr){2-5} \cmidrule(lr){6-9}
& \multicolumn{2}{c}{\cellcolor{yellow}LLM-as-Judge} & \multicolumn{2}{c|}{\cellcolor{lightblue}Automatic Metrics} & \multicolumn{2}{c}{\cellcolor{yellow}LLM-as-Judge} & \multicolumn{2}{c}{\cellcolor{lightblue}Automatic Metrics} \\
\cmidrule(lr){2-3} \cmidrule(lr){4-5} \cmidrule(lr){6-7} \cmidrule(lr){8-9}
& Correct & Complete & BERTScore & ROUGE-L & Correct & Complete & BERTScore & ROUGE-L \\
\midrule
\rowcolor{gray!10}\multicolumn{9}{l}{\textbf{Existing Medical Finetuned Model}}  \\
\addlinespace[2pt]
Meditron-7B\textsuperscript{*}  & 3.63  {\scriptsize $\pm$  2.15} & 1.32 {\scriptsize$\pm$ 1.16} & 60.60 {\scriptsize$\pm$ 1.20} & 3.10 {\scriptsize$\pm$ 0.50} & 2.99 {\scriptsize$\pm$ 1.62} & 1.00 {\scriptsize$\pm$ 0.87} & 65.23 {\scriptsize$\pm$ 1.00} & 5.39 {\scriptsize$\pm$ 0.61} \\
MedAlpaca\textsuperscript{*} & 12.87 {\scriptsize$\pm$ 3.63} & 4.29 {\scriptsize$\pm$ 2.15} & 65.90 {\scriptsize$\pm$ 1.10} & 4.80 {\scriptsize$\pm$ 0.90} & 7.21 {\scriptsize$\pm$ 2.49} & 1.49 {\scriptsize$\pm$ 1.12} & 72.06 {\scriptsize$\pm$ 0.74} & 9.25 {\scriptsize$\pm$ 0.79} \\
AlpaCare\textsuperscript{*} & 27.72 {\scriptsize$\pm$ 4.95} & 12.87 {\scriptsize$\pm$ 3.80} & 66.50 {\scriptsize$\pm$ 2.70} & 11.30 {\scriptsize$\pm$ 1.20} & 7.71 {\scriptsize$\pm$ 2.61} & 1.24 {\scriptsize$\pm$ 1.12} & 75.07 {\scriptsize$\pm$ 0.10} & 14.39 {\scriptsize$\pm$ 0.75} \\
MMed-LLaMA-3-8B\textsuperscript{*}  & 9.24 {\scriptsize$\pm$ 3.30} & 4.29 {\scriptsize$\pm$ 2.15} & 65.60 {\scriptsize$\pm$ 0.80} & 4.90 {\scriptsize$\pm$ 0.60} & 17.66 {\scriptsize$\pm$ 3.73} & 6.72 {\scriptsize$\pm$ 2.36} & 72.77 {\scriptsize$\pm$ 0.55} & 10.80 {\scriptsize$\pm$ 0.60} \\
PMC-LLaMA-13B\textsuperscript{*} & 11.88 {\scriptsize$\pm$ 3.63} & 4.62 {\scriptsize$\pm$ 2.48} & 65.00 {\scriptsize$\pm$ 1.10} & 3.70 {\scriptsize$\pm$ 0.60} & 1.24 {\scriptsize$\pm$ 1.12} & 0.50 {\scriptsize$\pm$ 0.62} & 29.17 {\scriptsize$\pm$ 2.81} & 0.77 {\scriptsize$\pm$ 0.35} \\
MedLM-Large\textsuperscript{*}\textsuperscript{†}  & 41.30 {\scriptsize$\pm$ 5.60} & 20.80 {\scriptsize$\pm$ 4.60} & 75.86 {\scriptsize$\pm$ 0.99} & 13.78 {\scriptsize$\pm$ 1.27} & 22.44 {\scriptsize$\pm$ 4.11} & 8.73 {\scriptsize$\pm$ 2.75} & 82.19 {\scriptsize$\pm$ 0.44} & 22.43 {\scriptsize$\pm$ 1.16} \\
MedLM-Medium\textsuperscript{†}  & 50.50 {\scriptsize$\pm$ 5.60} & 29.40 {\scriptsize$\pm$ 5.30} & 75.61 {\scriptsize$\pm$ 0.90} & 13.21 {\scriptsize$\pm$ 1.34} & 47.76 {\scriptsize$\pm$ 4.73} & 22.64 {\scriptsize$\pm$ 4.10} & 83.27 {\scriptsize$\pm$ 0.47} & 24.33 {\scriptsize$\pm$ 1.26} \\
MedInstruct\textsuperscript{‡}  & 45.90 {\scriptsize$\pm$ 5.60} & 27.70 {\scriptsize$\pm$ 5.10} & 70.90 {\scriptsize$\pm$ 0.70} & 8.70 {\scriptsize$\pm$ 0.60} & 59.45 {\scriptsize$\pm$ 4.73} & 38.81 {\scriptsize$\pm$ 4.73} & 80.14 {\scriptsize$\pm$ 0.47} & 18.86 {\scriptsize$\pm$ 0.73} \\
\midrule
\rowcolor{gray!10}\multicolumn{9}{l}{\textbf{\textsc{\instruct} Tuned Model with Different Base}}  \\
\addlinespace[2pt]
Qwen2.5-7B-Instruct  & 58.42 {\scriptsize$\pm$ 5.61} & 41.91 {\scriptsize$\pm$ 5.61} & 73.57 {\scriptsize$\pm$ 0.54} & 9.51 {\scriptsize$\pm$ 0.62} & 67.41{\scriptsize$\pm$4.48} & 53.48{\scriptsize$\pm$4.85} & 80.81{\scriptsize$\pm$0.34} & 18.40{\scriptsize$\pm$0.64} \\
\rowcolor{green!10}w/ \textsc{\instruct} Tuning  & 60.40 {\scriptsize$\pm$ 5.61} & 43.23 {\scriptsize$\pm$ 5.61} & 73.39 {\scriptsize$\pm$ 0.54} & 9.36 {\scriptsize$\pm$ 0.62} & 69.15{\scriptsize$\pm$4.48} & 52.99{\scriptsize$\pm$4.85} & 81.53{\scriptsize$\pm$0.35} & 19.12{\scriptsize$\pm$0.66} \\
Llama3.1-8B-Instruct  & 46.53 {\scriptsize$\pm$ 5.45} & 29.70 {\scriptsize$\pm$ 5.12} & 70.50 {\scriptsize$\pm$ 0.70} & 8.50 {\scriptsize$\pm$ 0.70} & 57.96 {\scriptsize$\pm$ 4.85} & 34.58 {\scriptsize$\pm$ 0.46} & 79.49 {\scriptsize$\pm$ 0.45} & 17.78 {\scriptsize$\pm$ 0.70} \\
\rowcolor{green!10} w/ \textsc{\instruct} Tuning  & 53.47 {\scriptsize$\pm$ 5.61} & 37.29 {\scriptsize$\pm$ 5.61} & 76.70 {\scriptsize$\pm$ 0.80} & 14.60 {\scriptsize$\pm$ 1.40} & 64.68 {\scriptsize$\pm$ 4.73} & 46.27 {\scriptsize$\pm$ 4.85} & 83.20 {\scriptsize$\pm$ 0.40} & 22.60 {\scriptsize$\pm$ 1.07} \\
\bottomrule
\multicolumn{9}{l}{\small 
\textsuperscript{*}These models have a maximum context length $\leq$ 8K. We truncated the most recent records to fit within their maximum size.} \\
\multicolumn{9}{l}{\small 
\textsuperscript{†}MedLM are powered by Med-PaLM 2, which is a medical fine-tuned version of Google PaLM.} \\
\multicolumn{9}{l}{\small 
\textsuperscript{‡}We instruct-tuned MedInstruct w/ Llama3.1-8B-Instruct as the base model.}
\end{tabular}
}
\vskip -1.5em
\end{table*}

\subsection{Experiment Setup} 
\textbf{Datasets.} \benchmark and \instruct utilize patient data from an academic medical center's research data repository, which contains records from its associated healthcare system, including both adult and children's hospitals. The repository follows the OMOP-CDM (Observational Medical Outcomes Partnership Common Data Model)~\cite{OHDSI_Data_Standardization} structure and encompasses 3.67M unique patients with records spanning from 1990 to February 8th, 2023. As all data was deidentified, this study did not require the approval of the Institutional Review Board. We preprocessed patient timelines into chunks that fit the instruction-tuning model's context window. Using Gemini-1.5-Pro~\cite{team2024gemini}, we generated 5000 instruction response pairs with temporal evidence supporting the generated answer. The prompt for instruction-tuning data generation can be found in Appendix~\ref{app:instruct_generation_prompt}. 
For \benchmark, we sampled separate patient timelines with no overlap with the instruction-tuning set. We filtered the benchmark questions by selecting those that have multiple time-stamped pieces of evidence, indicating that they require synthesis from different parts of the patient record. The prompt for benchmark generation can be found in Appendix~\ref{app:benchmark_generation_prompt}. Three clinicians validated each question's clinical relevance, temporal reasoning, and accuracy, which is detailed in Appendix~\ref{app:benchmark_clinician_verification}. 

As shown in Table~\ref{tab:benchmark-stats}, compared to existing benchmarks, \benchmark provides a more comprehensive view that incorporates both structured data and clinical notes across multiple visits with an average time span of 1,294.88 days while maintaining scalability through model-synthetic curation. Notably, \benchmark is the only dataset that explicitly includes temporal evidence annotations, facilitating more rigorous evaluation of temporal reasoning capabilities. 


\textbf{Model parameters.} We instruction-tuned Llama-3.1-8B-Instruct~\cite{llama3, hu2022lora} (8B parameters, 16K context) using LoRA. We perform a hyperparameter search to train the model and use the best hyperparameters as identified in Table~\ref{tab:hyperparameters}. We train for 6 epochs over a dataset of 5000 examples, using an effective batch size of 16. More details can be referred to in Appendix~\ref{app:hyper-parameter}. We further instruction-tune Qwen-2.5-7B-Instruct ~\cite{qwen2.5} to show the generalization of synthetic instruction tuning to improve model performance on both physician-generated and synthetically generated benchmarks.

\textbf{Evaluation metrics.} We evaluate models' open-text responses using LLM-Judge that assesses response correctness and completeness\footnote{Prompts detailed in Appendix~\ref{app:llm_judge_prompt}}, which is verified through correlation analysis with clinician evaluations, demonstrating strong alignment with human judgment with  $|\rho_{corr}|$ = 0.94 for correctness and $|\rho_{corr}|$ = 0.89 for completeness, as shown in Appendix~\ref{app:verify-llm-judge}. We also employ head-to-head comparisons and automated metrics derived from token-level representations, including BertScore \cite{zhang2019bertscore} (using distilbert-based-uncased), ROUGE-L \cite{lin2004rouge}, CHRF \cite{popovic2015chrf}, and METEOR \cite{banerjee2005meteor}\footnote{Results for these additional metrics are in Appendix~\ref{app:additional_metrics}.} to provide standard assessment of response quality. All LLM-based evaluations use GPT-4o-mini as the judge.
 
\vskip -0.5em
\begin{table}[t]
\caption{Head-to-head comparison between various models and \instruct. Each row shows the winning margin (additional wins by \instruct) and tie rates for both benchmarks.}
\label{tab:head2head}
\centering
\small  
\setlength{\tabcolsep}{3pt}  
\resizebox{1.0\linewidth}{!}{
\begin{tabular}{l|rr|rr}
\toprule
& \multicolumn{2}{c|}{\cellcolor{gray!15}{\textbf{MedAlign}}} & \multicolumn{2}{c}{\cellcolor{gray!15}{\textbf{\benchmark}}} \\
\textbf{Model} & \multicolumn{1}{c}{Win\%} & \multicolumn{1}{c|}{Tie\%} & \multicolumn{1}{c}{Win\%} & \multicolumn{1}{c}{Tie\%} \\
\midrule
\rowcolor{gray!15}\multicolumn{5}{l}{\emph{Existing Medical Finetuned Models}} \\
Meditron-7B & \cellcolor{green!5}+83.10 & 2.30 & \cellcolor{green!5}+95.02 & 0.50 \\
MedAlpaca & \cellcolor{green!5}+72.80 & 2.80 & \cellcolor{green!5}+86.41 & 0.37 \\
AlpaCare & \cellcolor{green!5}+84.40 & 1.00 & \cellcolor{green!5}+73.82 & 0.01 \\
MMed-LlaMa-3-8B & \cellcolor{green!5}+80.14 & 2.60 & \cellcolor{green!5}+81.71 & 0.87 \\
PMC-LlaMa-13B & \cellcolor{green!5}+74.40 & 4.10 & \cellcolor{green!5}+96.14 & 0.12 \\
MedLM-Large & \cellcolor{green!5}+27.80 & 6.60 & \cellcolor{green!5}+52.49 & 1.00 \\
MedLM-Medium & \cellcolor{green!5}+39.50 & 5.90 & \cellcolor{green!5}+20.65 & 1.99 \\
\midrule
\rowcolor{gray!15}\multicolumn{5}{l}{\emph{Models with the Same Base}} \\
MedInstruct (w/ Llama3.1-8B-Instruct) & \cellcolor{green!5}+6.30 & 6.90 & \cellcolor{green!5}+8.45 & 1.99 \\
Llama3.1-8B-Instruct & \cellcolor{green!5}+23.80 & 5.60 & \cellcolor{green!5}+17.67 & 1.99 \\
\bottomrule
\end{tabular}
}
\vskip -1.5em
\end{table}

\begin{table*}[t]
\centering
\caption{Performance analysis of instruction-tuning the same base models with different temporal distribution of instructions: recency-focused ($p_i > 0.75$), edge-focused (higher density at $t_{min}$ and $t_{max}$), and uniform distribution. We show head-to-head model comparisons, with \textbf{Model B} having an \underline{aligned temporal distribution} with the benchmark, and individual model metrics for each benchmark. The three benchmarks represent different temporal distributions: MedAlign (Recency), TIMER-Bench (Edge) with natural model-generated distribution, and TIMER-Bench (Uniform) with balanced temporal coverage. Bold numbers indicate the best performance in comparison.}
\label{tab:distribution-influence}
\resizebox{\linewidth}{!}{%
\begin{tabular}{l|ccc|ccc}
\toprule
& \multicolumn{3}{c|}{\textbf{Head-to-Head Comparison}} & \multicolumn{3}{c}{\textbf{LLM-as-Judge Metrics}} \\
\textbf{Benchmark} & Model A vs. Model B & Win Rate (A / B \%) & Tie (\%) & Model & Correctness & Completeness \\
\midrule
\multirow{3}{*}{\begin{tabular}[c]{@{}l@{}}\textcolor{medalign}{\textbf{MedAlign (Recency)}}\\\textcolor{gray}{\textit{Human-Annotated, Recency-Focused Distribution}}\end{tabular}} 
& \multirow{3}{*}[-0.3ex]{\begin{tabular}[c]{@{}c@{}}\textcolor{timer-bench-edge}{Edge}  vs. \textcolor{medalign}{\textbf{Recency}} \\[1ex]\textcolor{timer-bench-uniform}{Uniform}  vs. \textcolor{medalign}{\textbf{Recency}} \end{tabular}} & 
\multirow{3}{*}[-0.3ex]{\begin{tabular}[c]{@{}c@{}}42.24 / \textbf{43.89}\\[1ex]41.42 / \textbf{42.57}\end{tabular}} & 
\multirow{3}{*}[-0.3ex]{\begin{tabular}[c]{@{}c@{}}13.86\\[1ex]16.01\end{tabular}} & 
\textcolor{medalign}{\textbf{Recency}}  & \textbf{55.54} & 34.32 \\
& & & & \textcolor{timer-bench-edge}{Edge}  & 53.47 & \textbf{37.29} \\
& & & & \textcolor{timer-bench-uniform}{Uniform}  & 50.83 & 33.70 \\
\midrule
\multirow{3}{*}{\begin{tabular}[c]{@{}l@{}}\textcolor{timer-bench-edge}{\textbf{\benchmark (Edge)}}\\\textcolor{gray}{\textit{Model-Generated, Higher Density at $T_{min}$ and $T_{max}$}}\\\end{tabular}} 
& \multirow{3}{*}[-0.3ex]{\begin{tabular}[c]{@{}c@{}}\textcolor{medalign}{Recency}  vs. \textcolor{timer-bench-edge}{\textbf{Edge}} \\[1ex]\textcolor{timer-bench-uniform}{Uniform}  vs. \textcolor{timer-bench-edge}{\textbf{Edge}} \end{tabular}} & 
\multirow{3}{*}[-0.3ex]{\begin{tabular}[c]{@{}c@{}}47.89 / \textbf{48.76}\\[1ex]47.26 / \textbf{47.64}\end{tabular}} & 
\multirow{3}{*}[-0.3ex]{\begin{tabular}[c]{@{}c@{}}3.36\\[1ex]5.10\end{tabular}} & 
\textcolor{medalign}{Recency}  & 63.93 & 40.55 \\
& & & & \textcolor{timer-bench-edge}{\textbf{Edge}}  & 64.68 & \textbf{46.27} \\
& & & & \textcolor{timer-bench-uniform}{Uniform}  & \textbf{65.17} & 42.54 \\
\midrule
\multirow{3}{*}{\begin{tabular}[c]{@{}l@{}}\textcolor{timer-bench-uniform}{\textbf{\benchmark (Uniform)}}\\\textcolor{gray}{\textit{Model-Generated, Balanced Distribution on Timelines}}\end{tabular}} 
& \multirow{3}{*}[-0.3ex]{\begin{tabular}[c]{@{}c@{}}\textcolor{medalign}{Recency}  vs. \textcolor{timer-bench-uniform}{\textbf{Uniform}} \\[1ex]\textcolor{timer-bench-edge}{Edge}  vs. \textcolor{timer-bench-uniform}{\textbf{Uniform}} \end{tabular}} & 
\multirow{3}{*}[-0.3ex]{\begin{tabular}[c]{@{}c@{}}45.16 / \textbf{51.01}\\[1ex]47.58 / \textbf{48.39}\end{tabular}} & 
\multirow{3}{*}[-0.3ex]{\begin{tabular}[c]{@{}c@{}}3.83\\[1ex]4.03\end{tabular}} & 
\textcolor{medalign}{Recency}  & 63.71 & 39.92 \\
& & & & \textcolor{timer-bench-edge}{Edge}  & 61.69 & 43.55 \\
& & & & \textcolor{timer-bench-uniform}{\textbf{Uniform}}  & \textbf{64.52} & \textbf{43.55}\\
\bottomrule
\end{tabular}
}
\vskip -0.8em
\end{table*}

\subsection{Model Evaluation and Analysis} 
\textbf{Baselines.} 
We evaluate our approach against several baselines. These include existing medical fine-tuned models from the literature: Meditron-7B~\cite{epfmedtrn} with a 2K context, MedAlpaca~\cite{han2023medalpaca} with a 2K context, AlpaCare~\cite{zhang2023alpacare} with a 4K context, MMed-LlaMa-3-8B~\cite{qiu2024building} with an 8K context, PMC-LlaMa-13B~\cite{wu2024pmc} with an 8K context, and MedLM-Large~\cite{google_medpalm2_modelref} with an 8K context alongside MedLM-Medium~\cite{google_medpalm2_modelref} with a 16K context. For those models unable to process the entire EHR within their context window, we truncated the most recent $K$ tokens to fit the maximum context size. 
To identify the effectiveness of conventional medical instruction tuning for long-context reasoning, we include MedInstruct as an additional baseline, applying MedInstruct QA tuning ~\cite{zhang2023alpacare} to the long-context capable Llama-3.1-8B-Instruct. evaluate the role of instruction tuning with \instruct by applying it to different base models. Specifically, we tune both Qwen2.5-7B-Instruct~\cite{qwen2.5} and Llama3.1-8B-Instruct with \instruct, comparing their performance before and after tuning. 
For further reference, we evaluate several proprietary models (GPT4-32k, GPT-4o, Claude 3.5 Sonnet) and report their performance in Appendix~\ref{app:proprietary_model}.

\textbf{Results.} The results are in Table~\ref{tab:instruct-performance} and Table~\ref{tab:head2head}. Our evaluation reveals three key findings. 
First, despite domain-specific training on various medical datasets (e.g., MedQA, medical papers, clinical text), existing medical models with limited context windows struggle with long EHR tasks - even the best performing MedLM-Large achieves only 41.3\% correctness on MedAlign. While long-context models like MedLM-Medium and Llama3.1-8B-Instruct perform better, there remains significant room for improvement in temporal reasoning capabilities.
Second, simply applying short-form medical instruction tuning, i.e., MedInstruct QA to a long-context model, shows minimal gains over the base Llama3.1-8B-Instruct, showing 1.5\% improvement in correctness on \benchmark and even hurts performance on MedAlign, suggesting that traditional medical QA instruction tuning alone is insufficient for complex temporal reasoning in EHRs.
Finally, our temporal-aware instruction tuning approach \instruct demonstrates consistent improvements across model architectures. With Llama3.1-8B-Instruct as the base model, \instruct improves MedAlign correctness from 46.53\% to 53.47\% and completeness from 29.70\% to 37.29\% (average improvement of 7.3\%), while enhancing TIMER-Bench performance from 57.96\% to 64.68\% in correctness and 34.58\% to 46.27\% in completeness (average improvement of 9.2\%). These improvements generalize to other architectures: for Qwen2.5-7B-Instruct, \instruct improves MedAlign correctness from 58.42\% to 60.40\% and completeness from 41.91\% to 43.23\%, while similarly enhancing TIMER-Bench performance with correctness and automatic metrics. These gains across architectures demonstrate that our temporal-aware instruction tuning approach robustly enhances temporal reasoning and EHR understanding across models.

\begin{table*}[t]
\centering
\vskip -0.3em
\caption{Case study on \benchmark: model responses between base model Llama-3.1-8B-Instruct and model tuned w/ \instruct. }
\label{tab:short_cases}
\scriptsize 
\setlength{\tabcolsep}{3pt}  
\renewcommand{\arraystretch}{0.9}
\resizebox{1.0\linewidth}{!}{
\begin{tabular}{@{}p{0.2\textwidth}|p{0.22\textwidth}|p{0.25\textwidth}|p{0.24\textwidth}@{}}
\toprule
\textbf{Instruction} & \textbf{Llama-3.1-8B-Instruct} & w/ \textbf{\instruct}  & \textbf{Comparison Analysis} \\
\midrule
Describe the trend in the patient's weight over the past year, and assess if their current medication or dosage requires adjustment. & ....The patient's weight has been increasing over the past year, with a significant increase from 61.9 kg to 68 kg between 06/13/2021 and 12/18/2021. & ...On 12/18/2021, the patient weighed 68 kg (150 lb), which was 49\% for his age. On 01/21/2023, the patient weighed 64 kg (141 lb)...the patient's weight has decreased over the past year... & \textit{\textcolor{mypink}{\textbf{Improved temporal boundary adherence}}: The base model does not assess the trend within the last year, instead assessing the trend from 2+ years prior.}\\ 
\midrule
Has the patient's mixed conductive and sensorineural hearing loss in the right ear shown any improvement over time? & ...The patient's mixed conductive and sensorineural hearing loss in the right ear has shown some improvement over time....  & ...There is no indication of improvement... & \textit{\textcolor{mypink}{\textbf{Enhanced trend accuracy}}: The base model incorrectly assumes improvement over time despite the record not showing improvement.}\\
\midrule
Describe the patient's growth progression from age 9 to 11, referencing specific height measurements and percentiles. & ...At age 9, the patient's height was 135.3 cm (05/13/2011), which was at the 67th percentile... & ...At age 9 (09/08/2011), the patient's height was 139 cm, which placed her at the 67th percentile... & \textit{\textcolor{mypink}{\textbf{Better temporal precision}}: The base model incorrectly states patient's height at age 9 as 135.3 cm, which is actually the height recorded at age 8}\\
\bottomrule
\end{tabular}
}
\vskip -1em
\end{table*}


\begin{figure}[t]
\includegraphics[width=\linewidth]{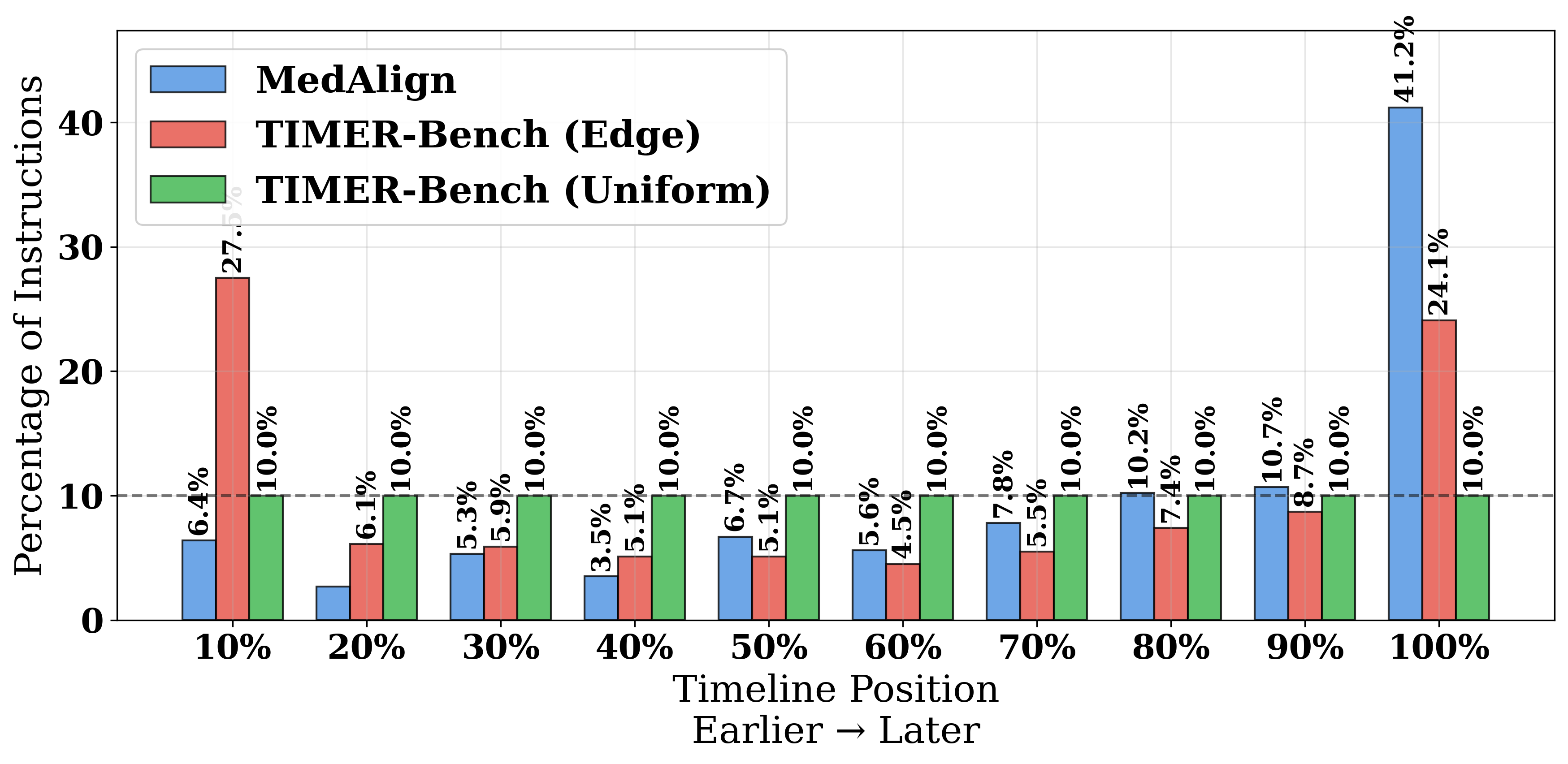} 
\vskip -1em
\caption{We evaluate on three benchmarks with varying temporal distributions: recent-focused, edge-focused, and uniform.}
\vskip -1em
\label{fig:benchmark-temporal-distribution}
\end{figure}

\subsection{Impact of Temporal Distribution Strategies}
\textbf{Settings.} Existing long-context EHR benchmarks primarily focus on questions about the most recent portions of patient timelines, creating an unintended distribution shift in evaluation. To enable a more comprehensive evaluation and isolate the impact of temporal reasoning from potential domain adaptation effects, we create \benchmark with two controlled temporal distributions: (1) an edge-focused distribution where evaluation instruction-response pairs are randomly sampled from the natural model-generated distribution and (2) a uniform distribution where the evaluation instruction-response pairs are sampled with equal frequency across all patient visits. As shown in Figure~\ref{fig:benchmark-temporal-distribution}, these constructed distributions provide complementary evaluation settings to MedAlign, where over half of the instructions focus on visits from the most recent quarter of patient histories.

\textbf{Results.} Table~\ref{tab:distribution-influence} analyzes how different temporal distributions in instruction-tuning affect model performance across various evaluation settings. Head-to-head comparisons reveal that the optimal instruction tuning distribution depends on the benchmark's temporal distribution.
On recency-focused MedAlign, the model trained with recency-focused instructions achieves optimal performance, achieving 55.54\% correctness and winning head-to-head comparisons against models trained with both temporally edge-focused and uniform instructions.
For edge-focused TIMER-Bench (Edge), all instruction distributions yield competitive performance, with edge-focused and uniform models showing similar performance. However, the uniform model achieves slightly higher correctness (65.17\% vs. 64.68\%), suggesting that comprehensive temporal coverage benefits even for edge-focused evaluation.
Most notably, on TIMER-Bench (Uniform), which requires reasoning across all temporal positions, the model tuned on uniformly distributed instructions demonstrates clear advantages. Models trained with uniform distribution substantially outperform recency-focused ones, achieving a win rate of 51.01\% vs 45.16\%. The uniform model also achieves the highest correctness (64.52\%) and completeness (43.55\%) scores, demonstrating the importance of balanced temporal coverage during instruction tuning for uniform instructions.
These results highlight two key insights: (1) the importance of aligning instruction-tuning temporal distributions with the intended evaluation distribution, and (2) instruct-tuning models with temporally uniform distributed instructions yield more robust performance across evaluation scenarios, particularly for tasks requiring reasoning over the full temporal range.

\subsection{Case Study}
Table \ref{tab:short_cases} presents comparative examples between the base model and \instruct on temporal reasoning tasks. The analysis reveals that \instruct demonstrates (1) improved temporal boundary adherence, where \instruct model correctly focuses on the specified time period (e.g., ``past year'') while the base model includes irrelevant historical data; (2) enhanced trend accuracy, where \instruct model correctly identifies patterns in patient measurements over time while the base model makes incorrect assumptions about improvement; and (3) better temporal precision, where the instruction-tuned model accurately matches specific dates with corresponding measurements while the base model confuses the chronological order of events. These cases demonstrate \instruct's effectiveness in improving temporal reasoning capabilities in longitudinal record analysis. Additionally, we observe that the instruction-tuned model provides more complete responses and focuses on relevant temporal information. More comprehensive examples can be found in Appendix~\ref{app:long_cases}.

\vskip -1em

\section{Conclusion and Discussion} 

This work reveals the ability to reason over temporal dependencies as a critical dimension in evaluating and instruction-tuning language models for clinical use. We uncover temporal biases in existing benchmarks that limit our understanding of model capabilities. To address this, we introduce a new temporal benchmark \benchmark, which explicitly includes time evidence and enables controlled evaluation across different temporal distributions. 
\instruct, our method for temporal instruction tuning, shows significant improvements on both physician-generated benchmarks and temporal reasoning benchmarks. 
While our experiments primarily focus on clinical records, the principles of temporal modeling apply to developing large language models in other fields that require reasoning over documents or sequences of events with complex temporal relationships. 
Our code is available at \href{https://anonymous.4open.science/r/TIMER-2874}{TIMER}, and \benchmark will be released under a research data use agreement upon approval.

\newpage
\section*{Impact Statement}
This work aims to advance the capabilities of language models in processing and understanding temporal information in longitudinal clinical records, which has significant implications for healthcare applications. 
The potential benefits include an enhanced ability to process longitudinal medical records, which could assist healthcare providers in better understanding patient histories, tracking disease progression, and identifying temporal patterns in medical conditions. This could lead to more informed clinical decision-making and improved patient care. Additionally, our methodology for temporal-aware instruction tuning could be adapted for other domains where processing longitudinally related information is crucial.
While our research demonstrates improved performance in temporal reasoning tasks, we acknowledge several important considerations regarding its deployment and impact. Our model is intended to serve as an assistive tool for healthcare professionals and should not be used as a standalone system for medical decision-making. The model's outputs should always be verified by qualified medical professionals, as errors in temporal reasoning could lead to incorrect understanding of patient histories. We also acknowledge potential biases in our training data and evaluation metrics, which might affect model performance across different demographic groups or medical conditions.
We encourage further research into robust evaluation methods for temporal reasoning in medical contexts and recommend careful consideration of deployment contexts to ensure appropriate use of this technology.

\newpage
\bibliographystyle{icml2025}
\balance
\bibliography{reference}
\newpage
\appendix
\onecolumn
\section{Prompt for \benchmark Generation}
\label{app:benchmark_generation_prompt}
\begin{figure*}[h]
\centering
  \includegraphics[width=\textwidth]{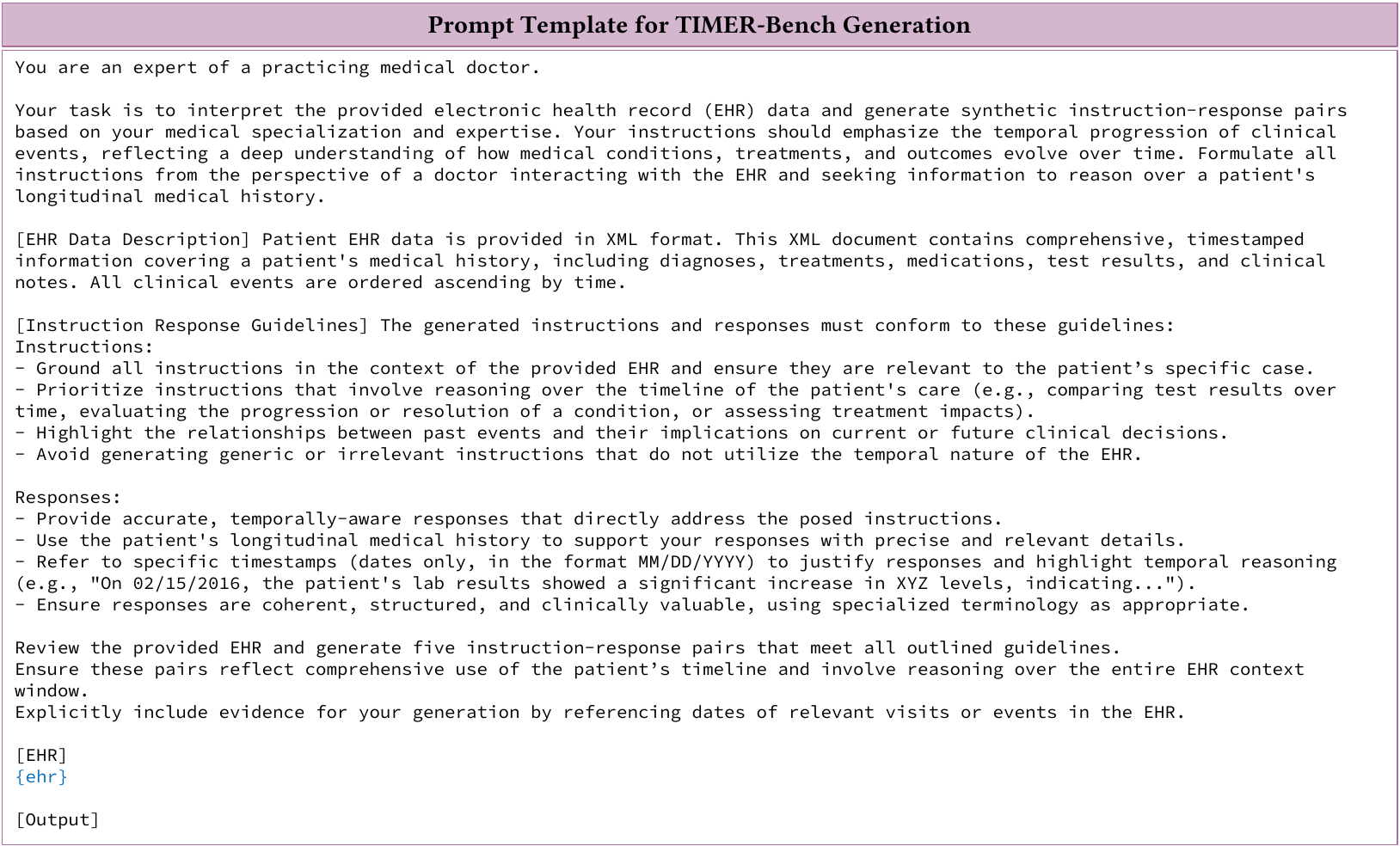}
    \vskip -1em
    \caption{Prompt template for \benchmark generation.}
    \vskip -1em
    \label{fig:prompt_bench_gen}
\end{figure*}
Figure \ref{fig:prompt_bench_gen} illustrates the prompt template used to generate TIMER-Bench, which guided Gemini-1.5-Pro to create temporal instruction-response pairs with time evidence from longitudinal records.

\section{\benchmark Clinician Verification}
\label{app:benchmark_clinician_verification}
We ask three clinicians to verify the relevance, quality, and accuracy of a randomly selected subset of the \benchmark instruction-response pairs. We note that the clinicians identified the generated instructions to have high relevance and the responses to be contextually accurate/reasonable, as they were not provided with the patient-protected information to do a complete factual accuracy evaluation. We also note that the majority of questions are considered to be medically complex by the clinicians, thus verifying that the synthetically generated questions are of reasonable complexity to capture model performance. The clinician-verified results are present in Table~\ref{tab:clinician_verification}.
\label{app:clinical-scores}
\begin{table}[h] 
    \centering
    \caption{Clinician Verification}
    \label{tab:clinician_verification}
    \resizebox{0.4\linewidth}{!}{%
    \begin{tabular}{@{}llll@{}}
        \toprule
        \textbf{Annotator} & \textbf{Clinical Relevance} & \textbf{Complexity} & \textbf{Accuracy}\\ \midrule
        Annotator 1         & 97/100              & 77/100 & 100/100           \\
        Annotator 2         & 89/100              & 63/100 & 96/100               \\         Annotator 3         & 99/100              & 99/100 & 100/100              \\ \bottomrule
    \end{tabular}
    }
\end{table}

\section{Prompt for Instruction-Tuning Data Generation}
\label{app:instruct_generation_prompt}
\begin{figure*}[h]
\centering
  \includegraphics[width=\textwidth]{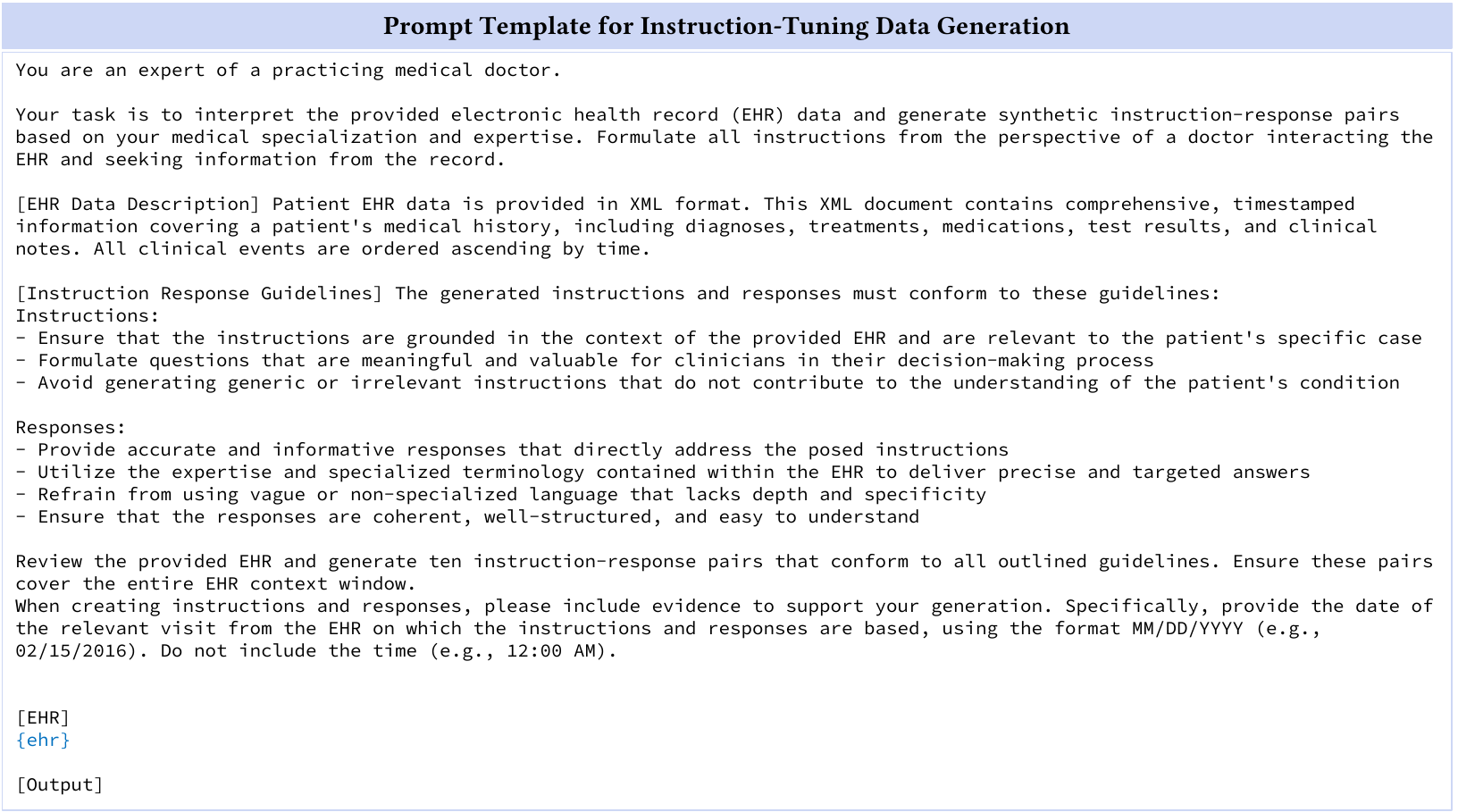}
    \vskip -1em
    \caption{Prompt template for instruction-tuning pairs generation.}
    \vskip -1em
    \label{fig:prompt_instruct_gen}
\end{figure*}
Figure \ref{fig:prompt_instruct_gen} shows the prompt template used to generate our instruction-tuning dataset, which guides Gemini-1.5-Pro in creating diverse instruction-response pairs that span different parts of patient records.

\section{Data Statistics}
\label{app:data_stats}
\begin{table}[h]
    \centering
    \caption{Lengths of questions and responses in each benchmark and instruction dataset.}
    \resizebox{0.6\linewidth}{!}{
    \begin{tabular}{l|c|ccc|ccc}
    \toprule
    \multirow{2}{*}{\textbf{Dataset}} & \multirow{2}{*}{Count} & \multicolumn{3}{c|}{\textbf{instructions}} & \multicolumn{3}{c}{\textbf{Responses}} \\
& & Q1 & Median & Q3 & Q1 & Median & Q3  \\
    \midrule
    MedAlign & 303 & 8 & 11 & 17 & 14 & 31 & 56 \\
    TIMER-Bench (Edge) & 402 & 14 & 18 & 22 & 42.25 & 59 & 82 \\
    TIMER-Bench (Uniform) & 248 & 14 & 18 & 21 & 42.75 & 61.5 & 82 \\
    \midrule
    TIMER-Instruct (Recency) & 5000 & 51 & 64 & 77 & 97 & 167 & 272 \\
    TIMER-Instruct (Edge) & 5000 & 52 & 64 & 77 & 100 & 167 & 259 \\
    TIMER-Instruct (Uniform) & 5000 & 54 & 65 & 78 & 103 & 168 & 259 \\
    \bottomrule
    \end{tabular}
    }
    \label{tab:data_stats}
\end{table}

Table \ref{tab:data_stats} shows the length distributions of instructions and responses across evaluation benchmarks, MedAlign and TIMER-Bench, and the three sets of instruction-tuning datasets, TIMER-Instruct.

\section{Hyper-parameter Tuning}
\label{app:hyper-parameter}
\begin{table}[h] 
    \centering
    \caption{Llama-3.1-8B-Instruct hyperparameter grid search.}
    \label{tab:hyperparameters}
    \resizebox{0.4\linewidth}{!}{%
    \begin{tabular}{@{}lll@{}}
        \toprule
        \textbf{Name} & \textbf{Values} & \textbf{Best Value} \\ \midrule
        Learning Rate         & 0, 5e-6, 1e-5, 1e-4              & 1e-5           \\
        Gradient Accumulation      & 4, 8, 16, 32                          & 16               \\         Weight Decay & 0, 1e-2, 1e-3,1e-4                          & 1e-4              \\ \bottomrule
    \end{tabular}
    }
\end{table}

Table \ref{tab:hyperparameters} presents the hyperparameter search space and optimal values for instruction-tuning Llama-3.1-8B-Instruct, where the best performance is achieved with a learning rate of 1e-5, gradient accumulation steps of 16, and weight decay of 1e-4.

\section{LLM Judge Prompt}
\label{app:llm_judge_prompt}
\begin{figure*}[h]
\centering
  \includegraphics[width=\textwidth]{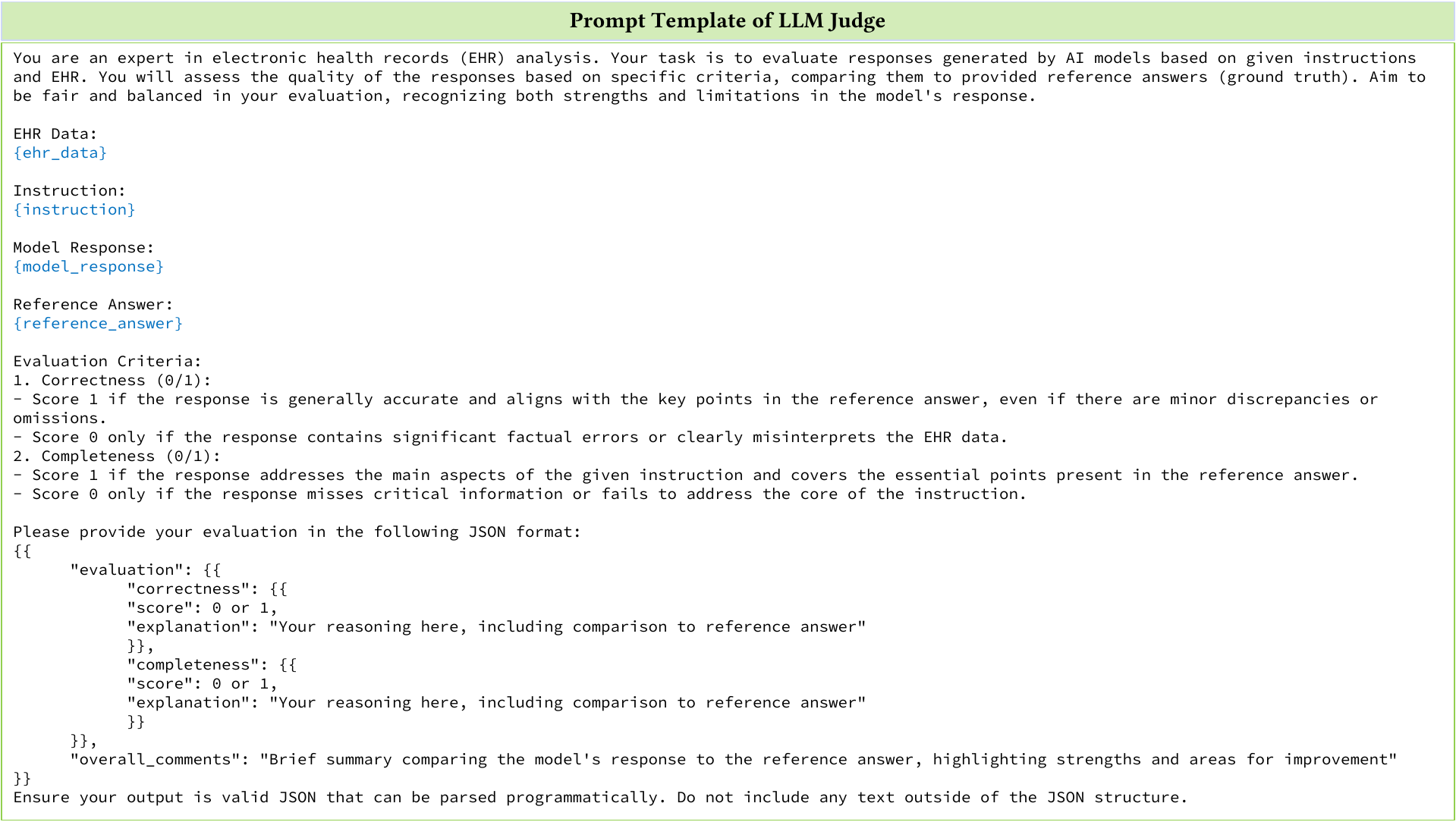}
    \vskip -1em
    \caption{Prompt template for LLM-as-Judge.}
    \vskip -1em
    \label{fig:prompt_llm_judge}
\end{figure*}
Figure \ref{fig:prompt_llm_judge} shows the prompt template used for the LLM-as-Judge evaluation, which we use to assess model outputs for correctness and completeness according to reference responses.

\section{The Correlation of LLM-Judge with Human Annotated Model Rank}
\label{app:verify-llm-judge}

\begin{table}[h]
\caption{Correlation between LLM-Judge evaluation and human judgments. Higher LLM scores and lower human ranks indicate better performance. Spearman correlation shows strong agreement between LLM scores and human ranks ($\rho=-0.97$ for average score, $\rho=-0.94$ for correctness, $\rho=-0.89$ for completeness).}
\label{tab:judge-human-correlation}
\centering
\resizebox{0.6\linewidth}{!}{%
\begin{tabular}{l|cc|c|c}
\toprule
\multirow{2}{*}{\textbf{Model}} & \multicolumn{2}{c|}{\textbf{LLM Score}} & \textbf{Human} & \textbf{LLM} \\
& Correctness & Completeness & Rank$\downarrow$ & Rank$\downarrow$ \\
\midrule
GPT4-32k & 0.419 & 0.360 & 2.309 & 1 \\
GPT4-32k-Multi-Step & 0.383 & 0.365 & 2.292 & 2 \\
Vicuna-13B & 0.343 & 0.292 & 3.259 & 3 \\
Vicuna-7B & 0.318 & 0.299 & 3.304 & 4 \\
MPT-7B-instruct & 0.193 & 0.149 & 3.688 & 5 \\
\bottomrule
\end{tabular}
}
\end{table}

To validate our LLM judge, we performed a correlation analysis between the scores from the LLM Judge and the model rankings annotated by clinicians based on the instructions of the MedAlign benchmark, where human evaluators assigned rankings to different models' responses to the same instructions. As shown in Table~\ref{tab:judge-human-correlation}, the LLM-judge metrics demonstrate strong agreement with human judgments, achieving high negative correlations for both correctness ($\rho_{corr}=-0.94$) and completeness ($\rho_{corr}=-0.89$). This inverse relationship aligns with our scoring scheme, where higher LLM scores and lower human ranks indicate better performance. Specifically, the ranking of models by LLM-judge scores consistently matches human preferences—GPT-4 variants are ranked highest, followed by Vicuna models, and then MPT-7B. This strong correlation suggests that our LLM-based evaluation framework provides a reliable proxy for human judgment in assessing temporal reasoning capabilities, enabling us to conduct larger-scale evaluations efficiently.

\section{Head-to-head Comparison Prompt}
\label{app:head2head_prompt}
\begin{figure*}[h]
\centering
  \includegraphics[width=\textwidth]{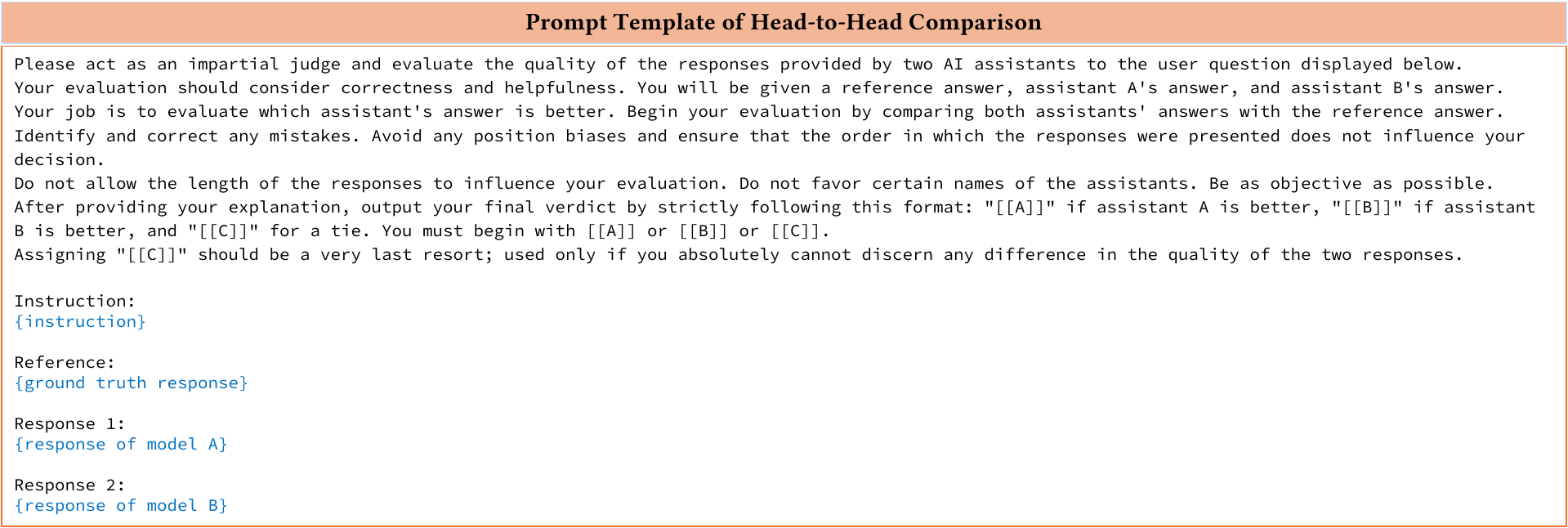}
    \vskip -1em
    \caption{Prompt template for head-to-head comparison.}
    \vskip -1em
    \label{fig:prompt_head2head}
\end{figure*}
Figure \ref{fig:prompt_head2head} displays the prompt template used for the head-to-head comparison between model outputs, where responses from different models are compared to evaluate their relative performance given the instruction and the reference answer.

\section{Proprietary Model Performance}
\label{app:proprietary_model}
\begin{table*}[h]
\centering
\caption{Proprietary model performance (\%) on MedAlign and \benchmark: reference as cap.}
\label{tab:proprietary-performance}
\resizebox{1.0\linewidth}{!}{
\begin{tabular}{lc|cccc|cccc}
\toprule
\multirow{4}{*}{\bf Model} & \multirow{4}{*}{\bf Open Source} & \multicolumn{4}{c|}{\bf MedAlign} & \multicolumn{4}{c}{\bf \benchmark} \\
\cmidrule(lr){3-6} \cmidrule(lr){7-10}
& & \multicolumn{2}{c}{\cellcolor{yellow}LLM-as-a-Judge} & \multicolumn{2}{c|}{\cellcolor{lightblue}NLP Metrics} & \multicolumn{2}{c}{\cellcolor{yellow}LLM-as-a-Judge} & \multicolumn{2}{c}{\cellcolor{lightblue}NLP Metrics} \\
\cmidrule(lr){3-4} \cmidrule(lr){5-6} \cmidrule(lr){7-8} \cmidrule(lr){9-10}
& & Correct.$\uparrow$ & Complete.$\uparrow$ & BERTScore$\uparrow$ & ROUGE-L$\uparrow$ & Correct.$\uparrow$ & Complete.$\uparrow$ & BERTScore$\uparrow$ & ROUGE-L$\uparrow$ \\
\midrule
\rowcolor{gray!15}\multicolumn{10}{l}{\emph{Proprietary Long Context Model Performance }}  \\
GPT4-32k &\xmark & 62.73$\pm$5.61 & 39.60$\pm$5.61 & 78.12$\pm$0.69 & 15.77$\pm$1.33 & 72.89$\pm$4.35 & 58.46$\pm$4.85 & 81.79$\pm$0.45 & 22.19$\pm$0.97 \\
GPT-4o &\xmark & 63.37$\pm$5.45 & 45.21$\pm$5.61 & 76.87$\pm$0.71 & 14.23$\pm$1.21 & 85.32$\pm$3.48 & 70.65$\pm$4.35 & 83.51$\pm$0.34 & 22.60$\pm$0.89 \\
Claude 3.5 Sonnet & \xmark & 68.98$\pm$5.28 & 57.76$\pm$5.45 & 77.17$\pm$0.55 & 12.21$\pm$0.86 & 89.55$\pm$2.99 & 88.31$\pm$3.11 & 81.41$\pm$0.29 & 16.91$\pm$0.57 \\
\bottomrule
\end{tabular}
}
\end{table*}
For reference purposes only, we present the performance of several proprietary large language models on both MedAlign and LongEHR-Bench datasets in Table~\ref{tab:proprietary-performance}. It is important to note that direct comparisons between these models and open-source alternatives would not be appropriate due to several factors: (1) the proprietary models have significantly larger model sizes and more extensive pretraining data, (2) their architectures and training procedures are not publicly available, and (3) their training costs are orders of magnitude higher than open-source alternatives. We include these results primarily to establish approximate performance ceilings for these benchmarks and to provide reference points for future research. The performance metrics are reported with standard deviations to account for the inherent variability in model outputs.

\section{Additional Metrics Results}
\label{app:additional_metrics}
\begin{table*}[t]
\centering
\caption{Additional automatic metrics results (\%) of baseline models and \instruct, reported as mean ± standard deviation from bootstrap resampling (n=10,000) over the test set with 100 samples. This table is over the entire MedAlign dataset.}
\label{tab:additional_metrics}
\resizebox{\linewidth}{!}{
\begin{tabular}{l|ccc|ccc|ccc}
\toprule
\multirow{2}{*}{\textbf{Model}} & \multicolumn{3}{c|}{\textbf{MedAlign}} & \multicolumn{3}{c|}{\textbf{TIMER-Bench (Edge)}}  & \multicolumn{3}{c}{\textbf{TIMER-Bench (Uniform)}} \\
\cmidrule(lr){2-4}\cmidrule(lr){5-7}\cmidrule(lr){8-10}
& METEOR & CHRF & Google BLEU & METEOR & CHRF & Google BLEU & METEOR & CHRF & Google BLEU \\
\midrule
\rowcolor{gray!10}\multicolumn{10}{l}{\textbf{Existing Medical Finetuned Model}}  \\
Meditron-7B\textsuperscript{*} & 6.22{\scriptsize$\pm$0.98} & 11.10{\scriptsize$\pm$0.80} & 1.19{\scriptsize$\pm$0.20} & 9.20{\scriptsize$\pm$1.14} & 16.64{\scriptsize$\pm$0.88} & 3.40{\scriptsize$\pm$0.31} & 8.23{\scriptsize$\pm$1.11} & 15.75{\scriptsize$\pm$1.07} & 3.24{\scriptsize$\pm$0.41} \\
MedAlpaca\textsuperscript{*} & 7.65{\scriptsize$\pm$1.04} & 11.30{\scriptsize$\pm$1.18} & 2.28{\scriptsize$\pm$0.60} & 12.81{\scriptsize$\pm$1.36} & 18.99{\scriptsize$\pm$0.98} & 5.52{\scriptsize$\pm$0.50} & 12.13{\scriptsize$\pm$1.36} & 18.50{\scriptsize$\pm$12.54} & 5.47{\scriptsize$\pm$0.67} \\
AlpaCare\textsuperscript{*} & 17.69{\scriptsize$\pm$1.13} & 20.39{\scriptsize$\pm$1.51} & 5.46{\scriptsize$\pm$0.70} & 20.41{\scriptsize$\pm$1.35} & 25.43{\scriptsize$\pm$1.06} & 7.93{\scriptsize$\pm$0.57} & 20.08{\scriptsize$\pm$1.35} & 25.04{\scriptsize$\pm$1.27} & 7.55{\scriptsize$\pm$0.65} \\
MMed-LLaMA-3-8B\textsuperscript{*} & 10.16{\scriptsize$\pm$1.00} & 13.21{\scriptsize$\pm$0.92} & 1.76{\scriptsize$\pm$0.30} & 19.99{\scriptsize$\pm$1.24} & 24.71{\scriptsize$\pm$0.91} & 5.69{\scriptsize$\pm$0.41} & 19.31{\scriptsize$\pm$1.20} & 24.36{\scriptsize$\pm$1.13} & 5.49{\scriptsize$\pm$0.51} \\
PMC-LLaMA-13B\textsuperscript{*} & 6.66{\scriptsize$\pm$0.43} & 8.84{\scriptsize$\pm$1.09} & 1.70{\scriptsize$\pm$0.30} & 1.04{\scriptsize$\pm$0.36} & 1.34{\scriptsize$\pm$0.52} & 0.51{\scriptsize$\pm$0.22} & 0.66{\scriptsize$\pm$0.33} & 0.87{\scriptsize$\pm$0.46} & 0.33{\scriptsize$\pm$0.19} \\

MedLM-Large\textsuperscript{*}\textsuperscript{†} & 19.45{\scriptsize$\pm$1.66} & 21.99{\scriptsize$\pm$1.456} & 6.66{\scriptsize$\pm$0.78} & 27.73{\scriptsize$\pm$1.55} & 32.08{\scriptsize$\pm$1.15} & 14.24{\scriptsize$\pm$0.94} & 34.68{\scriptsize$\pm$1.55} & 31.15{\scriptsize$\pm$1.401} & 13.50{\scriptsize$\pm$1.12} \\

MedLM-Medium\textsuperscript{†} & 17.84{\scriptsize$\pm$1.73} & 20.51{\scriptsize$\pm$1.59} & 6.60{\scriptsize$\pm$0.85} & 34.68{\scriptsize$\pm$1.44} & 38.48{\scriptsize$\pm$1.20} & 17.57{\scriptsize$\pm$1.07} & 35.18{\scriptsize$\pm$1.33} & 38.23{\scriptsize$\pm$1.41} & 16.75{\scriptsize$\pm$1.10} \\
MedInstruct\textsuperscript{‡} & 19.30{\scriptsize$\pm$1.20} & 20.66{\scriptsize$\pm$1.11} & 3.60{\scriptsize$\pm$0.40} & 35.18{\scriptsize$\pm$1.33} & 38.11{\scriptsize$\pm$0.91} & 11.89{\scriptsize$\pm$0.68} & 37.76{\scriptsize$\pm$1.32} & 37.30{\scriptsize$\pm$1.15} & 11.73{\scriptsize$\pm$0.68} \\
\midrule
\rowcolor{gray!10}\multicolumn{10}{l}{\textbf{\textsc{\instruct} Tuned Model with Different Base}} \\
Qwen2.5-7B-Instruct & 21.43{\scriptsize$\pm$1.15}&22.23{\scriptsize$\pm$1.05} & 4.11{\scriptsize$\pm$0.38}& 36.24{\scriptsize$\pm$0.94} & 38.96{\scriptsize$\pm$0.79} & 11.76{\scriptsize$\pm$0.54} & 35.51{\scriptsize$\pm$1.22} & 38.45{\scriptsize$\pm$1.01} & 11.62{\scriptsize$\pm$0.71} \\
\rowcolor{green!10}
{w/ \textsc{\instruct} Tuning} & 21.64{\scriptsize$\pm$1.13} & 21.73{\scriptsize$\pm$1.08}& 4.06{\scriptsize$\pm$0.39}& 37.38{\scriptsize$\pm$0.98} & 39.84{\scriptsize$\pm$0.84} & 12.61{\scriptsize$\pm$0.58} & 36.35{\scriptsize$\pm$1.24} & 39.32{\scriptsize$\pm$1.07} & 12.53{\scriptsize$\pm$0.74} \\
Llama3.1-8B-Instruct & 17.89{\scriptsize$\pm$1.02} & 19.63{\scriptsize$\pm$1.12} & 3.26{\scriptsize$\pm$0.40} & 33.84{\scriptsize$\pm$1.02} & 36.77{\scriptsize$\pm$0.91} & 10.80{\scriptsize$\pm$0.57} & 33.64{\scriptsize$\pm$1.44} & 36.41{\scriptsize$\pm$1.18} & 10.76{\scriptsize$\pm$0.59} \\
\rowcolor{green!10}
{w/ \textsc{\instruct} Tuning} & {22.91{\scriptsize$\pm$1.04}} & {25.43{\scriptsize$\pm$1.57}} & {7.88{\scriptsize$\pm$1.00}} & {37.76{\scriptsize$\pm$1.32}} & {41.17{\scriptsize$\pm$1.06}} & {16.82{\scriptsize$\pm$0.96}} & {37.10{\scriptsize$\pm$1.10}} & {40.19{\scriptsize$\pm$1.29}} & {16.23{\scriptsize$\pm$1.10}} \\
\bottomrule
\multicolumn{10}{l}{\small\textsuperscript{*}These models have a maximum context length $\leq$ 8K. We truncated the most recent records to fit within their maximum size.} \\
\multicolumn{10}{l}{\small 
\textsuperscript{†}MedLM are powered by Med-PaLM 2, which is a medical fine-tuned version of Google PaLM.} \\
\multicolumn{10}{l}{\small 
\textsuperscript{‡}We instruct-tuned MedInstruct w/ Llama3.1-8B-Instruct as the base model.}
\end{tabular}
}
\end{table*}

Table \ref{tab:additional_metrics} presents additional automatic evaluation metrics (METEOR, CHRF, and Google BLEU) for our model and baselines across MedAlign and \benchmark benchmarks, where TIMER-Instruct consistently improves performance across all metrics compared to the base model and medical QA fine-tuning.

\section{Additional \benchmark Results}
\label{app:uniform_results}
\begin{table*}[h]
\centering
\caption{Performance (\%) of baseline models and \instruct on \benchmark (Uniform), reported as mean ± standard deviation from bootstrap resampling (n=10,000) over the test set with 100 samples.}
\label{tab:instruct-performance-uniform}
\renewcommand{\arraystretch}{0.86}
\resizebox{0.55\linewidth}{!}{
\begin{tabular}{l|cccc}
\toprule
\multirow{2}{*}{\textbf{Model}} & \multicolumn{4}{c}{\textbf{TIMER-Bench (Uniform)}} \\
\cmidrule(lr){2-5}
& \multicolumn{2}{c}{\cellcolor{yellow}LLM Judge Metrics} & \multicolumn{2}{c}{\cellcolor{lightblue}NLP Metrics} \\
\cmidrule(lr){2-3} \cmidrule(lr){4-5}
& Correct & Complete & BERTScore & ROUGE-L \\
\midrule
\rowcolor{gray!10}\multicolumn{5}{l}{\textbf{Medical Finetuned Model}}  \\
\addlinespace[2pt]
Meditron-7B\textsuperscript{*} & 3.23{\scriptsize$\pm$2.22} & 0.40{\scriptsize$\pm$0.60} & 64.38{\scriptsize$\pm$1.38} & 4.91{\scriptsize$\pm$0.75} \\
MedAlpaca\textsuperscript{*} & 6.85{\scriptsize$\pm$3.02} & 1.61{\scriptsize$\pm$1.41} & 71.49{\scriptsize$\pm$0.97} & 8.82{\scriptsize$\pm$1.03} \\
AlpaCare\textsuperscript{*} & 7.26{\scriptsize$\pm$3.23} & 1.21{\scriptsize$\pm$1.41} & 75.20{\scriptsize$\pm$1.21} & 13.89{\scriptsize$\pm$0.83} \\
MMed-LLaMA-3-8B\textsuperscript{*} & 17.74{\scriptsize$\pm$4.64} & 6.85{\scriptsize$\pm$3.02} & 72.84{\scriptsize$\pm$0.71} & 10.37{\scriptsize$\pm$0.75} \\
PMC-LLaMA-13B\textsuperscript{*} & 0.81{\scriptsize$\pm$1.01} & 0.40{\scriptsize$\pm$0.60} & 29.01{\scriptsize$\pm$3.44} & 0.44{\scriptsize$\pm$0.29} \\
MedLM-Large\textsuperscript{*}\textsuperscript{†} & 19.35{\scriptsize$\pm$4.84} & 6.05{\scriptsize$\pm$3.02} & 81.97{\scriptsize$\pm$0.56} & 21.24{\scriptsize$\pm$1.25} \\
MedLM-Medium\textsuperscript{†} & 48.79{\scriptsize$\pm$6.05} & 24.60{\scriptsize$\pm$5.44} & 82.99{\scriptsize$\pm$0.59} & 23.18{\scriptsize$\pm$1.41} \\
MedInstruct\textsuperscript{‡} & 58.47{\scriptsize$\pm$6.05} & 36.69{\scriptsize$\pm$6.05} & 79.93{\scriptsize$\pm$0.63} & 18.37{\scriptsize$\pm$0.94} \\
\midrule
\rowcolor{gray!10}\multicolumn{5}{l}{\textbf{\textsc{\instruct} Tuned Model with Different Base}}  \\
\addlinespace[2pt]
Qwen2.5-7B-Instruct & 64.52{\scriptsize$\pm$6.05} & 50.81{\scriptsize$\pm$6.05} & 80.65{\scriptsize$\pm$0.46} & 17.84{\scriptsize$\pm$0.77} \\
\rowcolor{green!10} w/ \textsc{\instruct} Tuning & 66.13{\scriptsize$\pm$5.85} & 47.58{\scriptsize$\pm$6.25} & 81.44{\scriptsize$\pm$0.45} & 18.55{\scriptsize$\pm$0.85} \\
Llama3.1-8B-Instruct & 58.47{\scriptsize$\pm$6.05} & 33.47{\scriptsize$\pm$6.05} & 79.30{\scriptsize$\pm$0.59} & 17.62{\scriptsize$\pm$0.95} \\
\rowcolor{green!10} w/ \textsc{\instruct} Tuning & {64.52}{\scriptsize$\pm$6.05} & {43.55}{\scriptsize$\pm$6.05} & {84.09}{\scriptsize$\pm$0.56} & {24.16}{\scriptsize$\pm$1.27} \\
\bottomrule
\multicolumn{5}{l}{\small \textsuperscript{*}Models with context length $\leq$ 8K; most recent records truncated to fit.} \\
\multicolumn{5}{l}{\small \textsuperscript{†}MedLM: medical fine-tuned version of Google PaLM (Med-PaLM 2).} \\
\multicolumn{5}{l}{\small \textsuperscript{‡}Instruct-tuned using Llama3.1-8B-Instruct as base model.}
\end{tabular}
}
\end{table*}

Table \ref{tab:instruct-performance-uniform} shows model performance on the uniformly distributed version of TIMER-Bench, where TIMER-Instruct demonstrates consistent advantages over both medical-finetuned baselines and general long-context models, achieving the best performance across all metrics.

\section{Case Studies}
\label{app:long_cases}
\begin{table*}[t]
\caption{Case studies (Part 1): Qualitative comparison of model responses between base Llama-3.1-8B-Instruct and \instruct-tuned model on \benchmark. Key differences in response are highlighted in \textcolor{spurple}{purple}, with analytical insights in \textcolor{spink}{pink}.}
\label{tab:case_studies1}
\small
\begin{tabular}{p{0.2\textwidth}|p{0.36\textwidth}|p{0.36\textwidth}}
\toprule
\textbf{Clinical Query} & \textbf{Llama-3.1-8B-Instruct} & \textbf{w/ \base-Instruct} \\
\midrule
\parbox[t]{0.2\textwidth}{\textbf{Case 1: Timeline Analysis}\\\small Query: The patient was seen on 01/16/2013 for hip and shoulder pain. Provide a timeline of the events leading to this visit, including relevant symptoms and their duration.} & 
\parbox[t]{0.36\textwidth}{\textbf{Response:} "The patient was seen on 01/16/2013 for hip and shoulder pain. The timeline of events leading to this visit is as follows:- \textcolor{spurple}{The patient was seen on 12/31/2011 for an ear lesion, which was described as a bump in the ear that was bleeding 2 days ago and had some crust over it.- The patient was seen on 06/26/2012 for a follow-up visit, where she presented with bumps spreading on her face, which were later diagnosed as molluscum contagiosum infection.- The patient was seen on 11/20/2012 for a hospital visit, but the note does not mention any symptoms related to hip or shoulder pain.- The patient was seen on 11/27/2012 for a hospital visit, but the note does not mention any symptoms related to hip or shoulder pain.}- The patient was seen on 01/09/2013 for a telephone encounter note, where her mother reported that the patient was experiencing asthma symptoms, including shoulder and chest pain when she runs or exercises.- The patient was seen on 01/16/2013 for a follow-up visit, where she presented with hip and shoulder pain, which she reported started after she"
\\\\\textit{Analysis: The model response inaccurately represents the timeline of events. \textcolor{spink}{It fails to mention the specific injury to the hip during PE that occurred a month prior to the visit on 01/16/2013.} Additionally, it does not accurately capture the duration and worsening of the shoulder pain and chest pain symptoms, which were reported as occurring over a couple of months leading up to the visit. The response \textcolor{spink}{gets distracted by visits in the patient record that do not correspond with the given instruction}.}} &
\parbox[t]{0.36\textwidth}{\textbf{Response:} \textcolor{spurple}{"The patient presented to the clinic on 01/16/2013 with complaints of hip pain and pinching pain in her left shoulder with running.} The hip pain had started a month prior to the visit, after she injured her hip in PE doing an obstacle course. The pain initially ignored, but worsened over time. The shoulder pain had been bothering her for about 2 months, specifically when she ran. She also experienced chest pain during exercise, which was alleviated with the use of albuterol. Additionally, she reported shortness of breath. The patient had previously experienced asthma symptoms, for which she had used inhalers in the past. She had also been prescribed Singular as a child. The patient was advised to rest for 2 weeks, use ibuprofen as needed, and resume activity with the use of an inhaler pre-exercise. She was also administered the flu mist vaccine. The patient was advised to call if her symptoms did not improve. The exact duration of the shoulder pain is not specified in the note. However, it is stated that it had been bothering her for \"maybe a couple of months\". The hip pain had been present for a month. The chest pain was only mentioned in relation."
\\\\\textit{Analysis: 
\textcolor{spink}{The model accurately captures the core facts regarding the patient's hip and shoulder pain, including the timeline of events leading to the visit on 01/16/2013.} It correctly identifies the injury during PE, the worsening of hip pain, and the shoulder pain experienced during running. The mention of asthma symptoms and the alleviation of chest pain with albuterol is also correctly noted.}} \\
\midrule
\parbox[t]{0.2\textwidth}{\textbf{Case 2: Medication Review}\\\small Query: The patient is concerned about his current narcotic use. What is his current regimen, and has he expressed concerns about this previously?} &
\parbox[t]{0.36\textwidth}{\textbf{Response:} "The patient is currently taking \textcolor{spurple}{methadone 60 mg per day, and narco 2 to 3 mg q.h.s}. He has expressed a concern about his narcotic use, stating that he has a fear of being on narcotics for the rest of his life and would like to be weaned off of his methadone. He has previously tried to take his medication differently than prescribed by his doctors while he was moving, because he was not able to stand up on his left leg much. He has also stated that he has tried to wean off of his methadone in the past, but was unable to do so. He has also stated that he has tried to wean off of his methadone in the past, but was unable to do so. He has also stated that he has ..."
\\\\\textit{Analysis: \textcolor{spink}{The model incorrectly states the patient's current narcotic regimen as Methadone 60 mg per day and Narco 2-3 mg q.h.s. }However, the reference answer specifies that the patient takes Methadone 60 mg daily, Narco 2-3 mg at bedtime, and Motrin 800 mg as needed. Additionally, the model response contains repetitive phrases about the patient's attempts to wean off Methadone.}} &
\parbox[t]{0.36\textwidth}{\textbf{Response:} "The patient is currently taking \textcolor{spurple}{methadone 60 mg per day, narco 2 to 3 mg q.h.s., and Motrin 800 mg}. He has been on a narcotic for some time. In the past, he had tried fentanyl patches, Vicodin, and morphine. He has expressed a concern about being on narcotics for the rest of his life, and would like to be weaned off of his methadone. He has also stated that he has had to increase his medication for a few days after moving, as he was unable to stand on his left leg much. He has been to another pain doctor in the past five years, Dr. Trout, and has also seen Dr. Galli. He is currently seeing Dr. Amling at the Stanford Pain Management Center. He is motivated to be weaned off of his methadone and tried on new medications."
\\\\\textit{Analysis:\textcolor{spink}{The model accurately states the patient's current medication regimen, including Methadone 60 mg daily, Narco 2-3 mg at bedtime, and Motrin 800 mg}. It correctly notes the patient's concerns about long-term narcotic use and his desire to be weaned off Methadone, aligning with the reference answer.}}  \\
\bottomrule
\end{tabular}
\end{table*}

\begin{table*}[t]
\caption{Case studies (Part 2): Qualitative comparison of model responses between base Llama-3.1-8B-Instruct and \instruct-tuned model on \benchmark. Key differences in response are highlighted in \textcolor{spurple}{purple}, with analytical insights in \textcolor{spink}{pink}.}
\label{tab:case_studies2}
\small
\begin{tabular}{p{0.18\textwidth}|p{0.37\textwidth}|p{0.37\textwidth}}
\toprule
\textbf{Clinical Query} & \textbf{Llama-3.1-8B-Instruct} & \textbf{w/ \base-Instruct} \\
\midrule
\parbox[t]{0.18\textwidth}{\textbf{Case 3: X-ray Findings Over Time}\\\small Query: The patient had a left knee injury in July 2022. Summarize the findings from his x-rays over time.} & 
\parbox[t]{0.37\textwidth}{\textbf{Response:} "The patient had a left knee injury in July 2022. Over time, the x-rays showed the following findings: 1. \textcolor{spurple}{Initially, there was a subtle irregularity along the proximal tibial metaphysis on lateral view X-ray on July 6, 2022, which raised concern for an occult proximal tibia fracture}. 2. Repeat x-rays on July 25, 2022, did not show any obvious bony abnormality, but there was soft tissue swelling in the area of tenderness. 3. Repeat x-rays on August 22, 2022, showed moderate sclerosis across the proximal tibial metaphysis as well as lateral periosteal reaction consistent with a healing nondisplaced fracture in this location. The joint spaces of the knee were preserved, and the soft tissues were radiographically normal. This confirmed the initial suspicion of a nondisplaced fracture of the proximal tibia. Answered correctly Answered correctly Answered correctly Answered correctly
Answered correctly"
\\\\\textit{Analysis: \textcolor{spink}{The model response contains factual inaccuracies regarding the timeline of x-ray findings. It incorrectly states that the initial x-ray was on July 6, 2022, while the correct date is July 18, 2022}. Additionally, it misrepresents the findings on July 25, 2022, as showing soft tissue swelling, which was not mentioned in the reference answer.}} &
\parbox[t]{0.37\textwidth}{\textbf{Response:} "1. \textcolor{spurple}{Initial x-rays on 07/18/2022} showed a subtle irregularity along the proximal tibial metaphysis on lateral view, raising concern for an occult proximal tibia fracture. 2. Repeat x-rays on 07/25/2022 showed no obvious bony abnormality, but there was soft tissue swelling in the area of tenderness. 3. Repeat x-rays on 08/22/2022 showed sclerosis across the proximal tibial metaphysis and a lateral periosteal reaction, consistent with a healing nondisplaced fracture in this location. The joint spaces of the knee were preserved, and the soft tissues were radiographically normal. 4. No further x-rays were mentioned in the provided EHR for the left knee after 08/22/2022. 5. The patient was advised to limit his weightbearing activity based on his symptoms, and follow-up was recommended in 2 weeks if he was not comfortably walking without a limp. 6. The patient was also advised to avoid high-risk activities such as going to the playground, bouncy houses, or trampolines. 7. Follow-up x-rays were recommended in 3 weeks if the patient had not returned to full running"
\\\\\textit{Analysis: The model accurately describes the findings from the x-rays over time, including the initial irregularity, the repeat x-ray findings, and the final assessment of the healing nondisplaced fracture. \textcolor{spink}{The temporal relationships are correctly represented, and the dates mentioned align with the EHR data.}}} \\
\midrule
\parbox[t]{0.18\textwidth}{\textbf{Case 4: Injury Follow-up}\\\small Query: The patient was seen on 02/13/2017 for a left ankle sprain. Summarize the treatment plan and outcome at his follow-up appointments.} &
\parbox[t]{0.37\textwidth}{\textbf{Response:} "\textcolor{spurple}{The patient was seen on 02/13/2017 for a left ankle sprain. At his follow-up appointment on 01/23/2017}, the patient was placed in a lace-up ankle brace for the next week. After a week of full-time wear, he was instructed to wean out of it at home, and if he remained pain-free, he could discontinue the brace. He was also advised to avoid physical education for the next week and to return to the clinic if he continued to favor the ankle after 3 weeks. At his follow-up appointment on 02/13/2017, the patient had improved significantly, with only a slight limp and no pain. The doctor decided that the patient did not require a prescription for physical therapy at that point and instructed him to continue to use the brace as needed. The patient was advised to follow up on an as-needed basis. The patient was seen on 02/14/2017 for a follow-up visit, but the note does not mention any new treatment plan or outcome. The patient was seen on 09/28/2017 for a gastroenterology visit, which is not related to his ankle injury..."
\\\\\textit{Analysis: The model response is ordered in such a way that it states that \textcolor{spink}{the patient was seen on 01/23/2017 for follow-up after it discusses the appointment on 02/13/2017, creating an incorrect temporal flow of events.} }} & 
\parbox[t]{0.37\textwidth}{\textbf{Response:} "\textcolor{spurple}{At the follow-up visit on 01/23/2017, the patient was placed in a lace-up ankle brace for the next week. After a week of full-time wear, the brace was to be taken off only to shower and sleep. The patient was to wean out of the brace at home and discontinue it if he remained pain-free. He was advised to avoid physical education for the next week and was told that if the limp persisted or he continued to favor the ankle after 3 weeks, he should return to the clinic. At the follow-up visit on 02/13/2017,} the patient's limp had improved, and he had full ankle range of motion without pain. He was no longer tender over the ATFL, CFL, or deltoid ligament. The plan was to refer him for physical therapy if the pain or limp persisted. The patient was advised to continue using the ankle brace as needed. The patient was able to return to physical education and was expected to get away from the brace. He was advised to follow up as needed."
\\\\\textit{Analysis: \textcolor{spink}{The model accurately describes the treatment plan and outcomes for the patient's follow-up appointments, including the transition from a short leg walking cast to a lace-up ankle brace and the patient's improvement in symptoms.} Its ordering of events is temporally grounded and clear, aligning with user preference.}} \\
\bottomrule
\end{tabular}
\end{table*}

As shown in Table~\ref{tab:case_studies1} and Table~\ref{tab:case_studies2}, Qualitative analysis of model outputs reveals key differences in temporal reasoning capabilities between the baseline model and \instruct model. The baseline model exhibits several characteristic failure modes: (1) recall interference, where it retrieves temporally irrelevant information from the context, e.g., listing unrelated historical visits in Case 1, (2) temporal ordering errors, where it misattributes dates and event sequences, e.g., incorrect x-ray dates in Case 3, and (3) repetitive generation patterns when handling long temporal sequences, e.g., redundant phrases about medication weaning in Case 2. In contrast, \instruct demonstrates improved temporal processing capabilities: it successfully filters relevant temporal information from the context window, maintains chronological consistency across multiple time points, and generates non-repetitive responses that accurately capture temporal relationships. When presented with complex queries requiring multi-step temporal reasoning. e.g., treatment progression in Case 4, \instruct shows robust performance in establishing causal relationships and maintaining contextual relevance throughout the generated response. These improvements suggest that our temporal-aware instruction tuning effectively enhances the model's ability to process and reason about time-dependent information in long-context scenarios.



\end{document}